\documentclass[11pt]{amsart}
\usepackage{amsfonts, amsmath, amssymb}
\usepackage[top=0.75in, bottom=0.75in, left=0.75in, right=0.75in]{geometry}
\usepackage{float}

\usepackage{xcolor}

\usepackage[T1]{fontenc}    % use 8-bit T1 fonts
\usepackage{hyperref}       % hyperlinks
\usepackage{url}            % simple URL typesetting
\usepackage{booktabs}       % professional-quality tables
\usepackage{amsfonts}       % blackboard math symbols
\usepackage{nicefrac}       % compact symbols for 1/2, etc.
\usepackage{microtype}      % microtypography
\usepackage{xcolor}         % colors
\usepackage{graphicx}
\usepackage{amsmath}
\usepackage{mathtools}
\usepackage{overpic}

\DeclareMathOperator{\ie}{\operatorname{i}}

\title{Varifold Moment Invariants\\ for Sustainable and Explainable\\ Contour Feature Extraction }

\author{G. Longari}
\address{Computer Vision Lab, Technische Universität Wien, Karlsplatz 13, 1040 Vienna, Austria \\
         Wolfgang Pauli Institut, Oskar-Morgensternplatz 1, 1090 Vienna, Austria}

\author{J.-C. \'Alvarez Paiva}
\address{U.M.R. CNRS 8524, U.F.R. de Math\'ematiques, 59655 Villeneuve d'Ascq C\'edex, France}

\author{A. B. Tumpach}
\address{Laboratoire Painlevé, Lille University, 59650 Villeneuve d’Ascq, France \\
         Wolfgang Pauli Institut, Oskar-Morgensternplatz 1, 1090 Vienna, Austria}
\begin{document}

\begin{abstract}
  We introduce Varifold Moments Invariants (VMI) as a unifying framework for many previously introduced Moment Invariants.
 These invariants are deeply related to other contour features that are invariant under translations and rotations, like Extended Gaussian Image, Elliptic Fourier Descriptors or Shape Distributions. The advantage of the varifold approach to moments consists in being able to combine the geometry of the region, its boundary, and the family of lines tangent to it, in order to create a substantial number of invariant features with high discriminating power and clear geometric meaning. By coupling our VMI feature extraction with the light feature classifiers Random Forest or Multi-Layer-Perceptron, we outperform  state-of-the-art approaches based on contours, while decreasing drastically the computational cost to the point of allowing our algorithm to run on light devices. We tested our approach on classification tasks on a large number of widely-used datasets of various types (leaves, objects, cells) and achieved high accuracy with a low number of geometrically interpretable features.
 \end{abstract}

\maketitle

\section{Introduction}
\begin{figure}[t]
    \centering
        \begin{overpic}[width=\linewidth,  ]{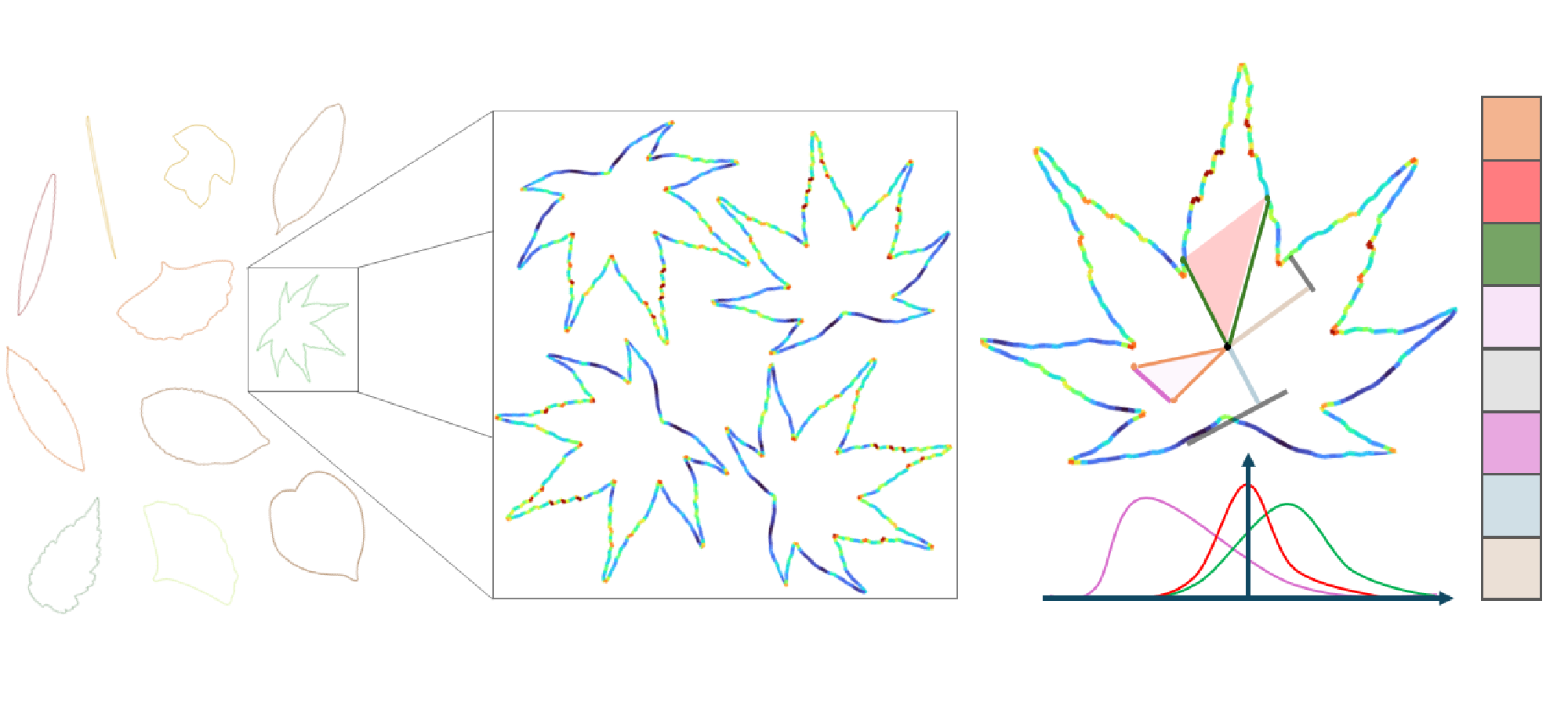} %grid, tics=10 

        \put(15, 4.5){\makebox(0,0){\small a)}}
        \put(15, 1){\makebox(0,0){\small 2D Shapes}}
        \put(48, 4.5){\makebox(0,0){\small b)}}
        \put(48, 2.2){\makebox(0,0){\small Rotation and}}
        \put(48, 0){\makebox(0,0){\small Translation invariants}}
        \put(80, 2.2){\makebox(0,0){\small Varifold}}
        \put(80, 0){\makebox(0,0){\small Moment Invariants}}
        \put(80, 4.5){\makebox(0,0){\small c)}}
        \put(96.5, 41.5){\makebox(0,0){\tiny VMI }}
        \put(96.5, 40){\makebox(0,0){\tiny Features}}
        %\put(73, 27.3){\makebox(0,0){\tiny $VMI_2$}}
        %\put(73, 23.6){\makebox(0,0){\tiny $VMI_3$}}
        %\put(73, 19.8){\makebox(0,0){\tiny $VMI_4$}}
        %\put(73, 16.0){\makebox(0,0){\tiny $VMI_5$}}
        %\put(73, 12.2){\makebox(0,0){\tiny $VMI_6$}}
        %\put(73, 8.5){\makebox(0,0){\tiny $VMI_i$}}
        %\put(73, 4.8){\makebox(0,0){\tiny $VMI_N$}}
        %\put(88,23){\makebox(0,0){\tiny $CM$}}
        %\put(87.5,22.5){\makebox(0,0){\tiny $VMI_1$}}

    \end{overpic}
    \caption{ Given 2D contours of objects a), we seek features that are independent of their position and orientation on the plane b). The Varifold Moment Invariants are extracted from the polygonal representation of the curve and capture different geometric features of the shape such as the moments of various shape distributions (e.g., signed areas of random triangles or distances from random tangent and normal lines to the center of mass), yielding robust representations for machine learning tasks.}
    \label{fig:abs}
\end{figure}
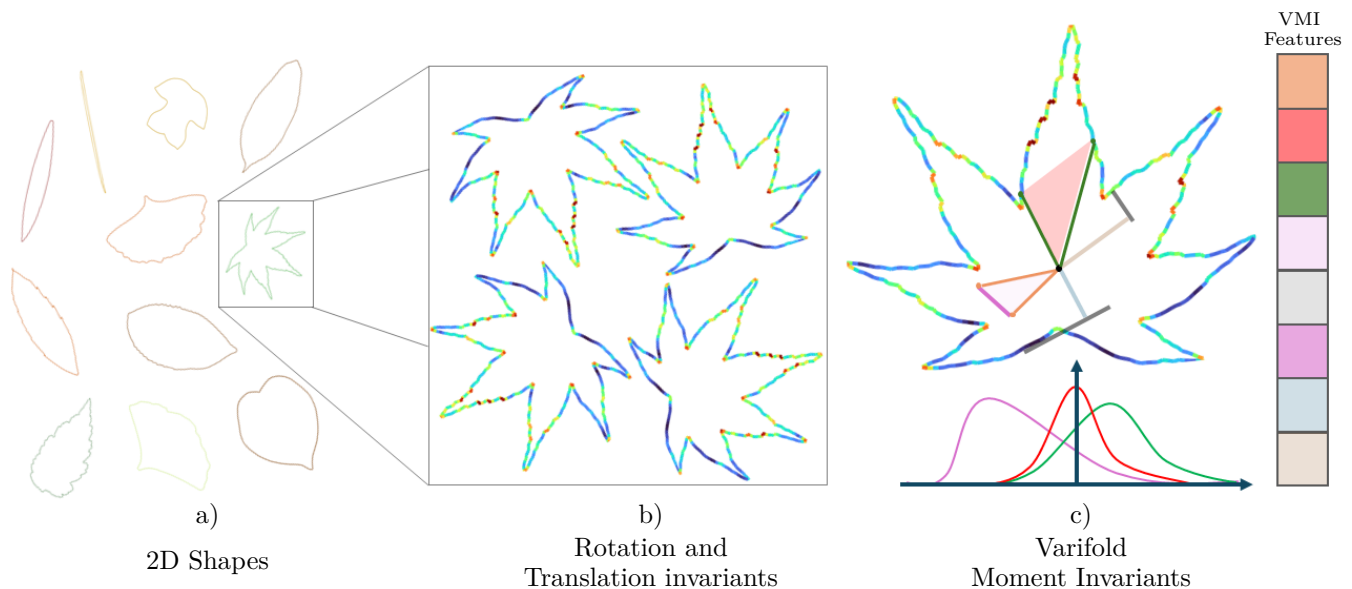

Varifolds \cite{Allard1972OnTF} are used in Computer Vision for shape registration and comparison \cite{Charon_Trouve, Kaltenmark, CHARON_Fidelity, LDDMM}, surface reconstruction and discrete approximation of surfaces \cite{Blanche_Buet_2015, varifold_approach}, partial matching \cite{Partial_Matching}, atlas estimation of Functional Shapes \cite{Charlier_Functional_shape}, shape interpolation \cite{Hartman2025, Bauer_Varifold_Curves}, comparison and classification of shape sequences \cite{Pierson_Human_Comparison, Pierson_thesis}, and have been generalized in \cite{Buet2023FlagfoldsAA}.

In the present work, we introduce Varifold Moments and Shape Invariants related to them, called \textit{Varifold Moment Invariants} (VMI). The present contribution leverages the theory of moments \cite{Hu1962, Abu-Mostafa_1984, FLUSSER_2000} to the varifold setting. We use this generalization  to design  high discriminating features of planar shapes bounded by simple closed curves modulo the action of a shape preserving group.
Depending on the applications, this is the group of reparameterizations, translations, and rotations,  which is sometimes extended by dilations.
Two curves will represent the same shape if there exists a group element that transforms one curve into the other.

 The invariants we introduce are products of \textit{Varifold Moments}, i.e. integrals of the form
 \begin{equation}\label{VM}
M(f;p,q,r) := \int_\gamma f(\kappa) \, z^p \overline{z}^q \left(\tfrac{\dot{z}}{|\dot{z}|}\right)^r |\dot{z}| \, dt,
\end{equation}
where $f$ is an arbitrary real-valued, piecewise-continuous function of the curvature $\kappa$, the powers $p, q \in \mathbb{N}$, $r\in\mathbb{Z}$, $z = x + \ie y$, and 
$|\dot{z}| \, dt$ the arc-length measure. These  contour integrals 
allow us to define new geometric invariants as well as to provide  a single framework for various classical moment invariants associated to different measures related to a curve (area measure of enclosed region, arc-length of the curve and extended Gaussian image). In particular our invariants encompass those in \cite{Hu1962, Abu-Mostafa_1984, FLUSSER_2000} and the Fourier coefficients of the extended Gaussian image \cite{Horn, persoon2007shape, Charon_Pierron, pierson2022projection}.

Many well-known Euclidean quantities associated to a curve $\gamma$ can be expressed using these contour integrals. For example, the length, area, total curvature, curvature centroid, and moments (with respect to both the arc-length measure of the curve and the area measure of the enclosed region) can all be written as $M(f;p,q,r)$ for appropriate choices of $f$ and integers $p, q, r$. Specifically, for a counter-clockwise oriented curve $\gamma$ bounding a domain $D$, the length and area are given by
\begin{equation}
\operatorname{Length}(\gamma) = M_{0,0,0} = \int_\gamma |\dot{z}| dt, \quad \quad \operatorname{Area}(\gamma) = \tfrac{1}{2\ie} M_{0,1,1} = \tfrac{1}{2i} \int_\gamma \overline{z} \, dt = \tfrac{1}{2i} \int_D d\overline{z} \wedge dz.
\end{equation}

Complex moments \eqref{complex_moments} can also be expressed as this type of contour integrals \cite{FLUSSER_Book}:
\begin{equation}\label{cpqMpqr}
c_{p,q}(D) = \iint_{D} z^p \bar{z}^q dxdy = \tfrac{1}{2i(q+1)} \int_{\partial D} z^p \overline{z}^{q+1} \, dt \\ 
= \tfrac{1}{2i(q+1)} M_{p,q+1,1}(\partial D).    
\end{equation}
We stress however that most varifold moments cannot be expressed in terms of their classical counterparts, even in the case where  the curvature is not taken into account. For instance, the invariant moment $M(1; 0,2,2)$ is not of this form. Geometrically, the classical moments associated to a region or  curve consider points as their basic geometric objects, the extended Gaussian image considers unit tangent vectors deprived of
their base points as the basic geometric objects, but in varifold moments the tangent 
unit vectors, base point and all, are considered. This allows us  to associate Varifold Moment Invariants to the geometry of tangent and normal lines.  

A key aspect of our approach that we found useful in feature design is that the moments of (1-dimensional) Shape distributions \cite{Osada} such as mutual squared distances between random points, signed areas of random triangles or signed distances from tangent lines to the center of mass, can be expressed in terms of Varifold Moments. We found the mean, variance, skewness and kurtosis of these distributions to be discriminating  Varifold Moment Invariants.

The main contributions of this paper are the following:
\begin{itemize}
    \item we propose a unified theory of Moments, called Varifold Moments, that encompass Hu's Moments \cite{Hu1962}, Harmonic moments \cite{DAVIS1977148}, Complex Moments \cite{Abu-Mostafa_1984, FLUSSER_2000}, Zernike Moments \cite{Wallin}, on curves and planar domains;
    \item we use Varifold Moments to construct features that are invariant under translations, rotations and/or scaling, called Varifold Moment Invariants (VMI), that form a complete list of shape descriptors;
    \item we use geometric insight to design new discriminating invariant features, like the mean, variance, skewness and kurtosis of geometric distributions related to the shape of contours, expressed as Varifold Moment Invariants, allowing interpretability of feature selection.
    \item Due to high Discriminability and Robustness of invariants constructed as combinations  of invariants for boundary measures, interior measures and Gaussian length measures, we propose algorithms with low environmental footprint.
\end{itemize}

\section{Related Work and Basic Definitions}

\subsection{Methods for invariance of Shape representations.}
\bigskip
\paragraph{\bf Extraction of invariant features.}
Features that are invariant or equivariant under translations and rotations can be extracted using  Region Property features \cite{van2014scikit}, Elliptic Fourier Descriptors \cite{kuhl1982elliptic, persoon2007shape}, Minkowski functionals \cite{Minkowski, Lines_E2_Equivariant_for_Radio_Galaxy}, Haralick features \cite{Haralick}, Measures along Chord Bunches \cite{CBW}, Representation Theory of Groups \cite{Weiler_unireps, sanborn_Miolane, Mataigne_Miolane_G_Bispectrum}, Invariant Moments \cite{Hu1962, FLUSSER_2000, Lukic, yuhheXShapeEnc2026, Anant_KNN_MomentINvariant_Texture, Jia_Leaf_Recognotion_Zernike_Moments}, Mutual Distances \cite{Kacem_Mutual_distances, romero2026shapeembed},  Gram-Matrices \cite{Kacem_Gram_Matrices}, Triangle Distance \cite{Yang_TDR}  \cite{Yang_TDR}, Multiscale Distance Matrices \cite{MDM}, integral invariant descriptors \cite{Integral_Invariants}, multiscale curvature-based shape representation \cite{Mokhtarian}, or Shape distributions \cite{Osada}. For domain specific classification, like leaves classification, multiple features extracted from shape, color and texture are usually combined \cite{Quach}. Reviews of state-of-the-art methods for Leaf classification  can be found in 
\cite{Wang_Review, ALgemayel_Review_ML_methods_Plants}, for cells in 
\cite{Vadori2024AutomatedCO}.
These methods can be coupled with machine learning algorithms like $K$-nearest-Neighbor \cite{KNN, Jia_Leaf_Recognotion_Zernike_Moments, Anant_KNN_MomentINvariant_Texture, Yang_TDR}, Support Vector Machine \cite{SVM, Lukic}, ResNet-18 \cite{ResNet, romero2026shapeembed},  LeNet \cite{LeNet, Lines_E2_Equivariant_for_Radio_Galaxy},  VGG16 \cite{VGG16, HuMomentforCovid19}. 
Since the features themselves remain invariant under the action of the shape-preserving group, the resulting algorithms achieve exact invariance.

\paragraph{\bf Forcing invariance in Learning Algorithms.}
When the features extracted are not invariant under the action of the shape preserving group, invariance  can be learned using data augmentation
in the sample space \cite{Dieleman}, data augmentation in the feature space \cite{Kumar_feature_augmentation}, prealignment  \cite{O2-VAE}
or modified convolutional neural networks \cite{Cohen_Group_Equivariant_CN, Finzi_Lie_Group_equivariance, Weiler_steerable_filters, Smets_PDE, Bekkers_Roto_Translation, gardaa2025rotatouillerotationequivariantdeep}. These methods usually suffer from high computational costs and are memory-intensive. Moreover, invariance is often only approximated, leading to the representation of shapes as multiple points in latent space (see Fig.2e in \cite{O2-VAE}).

\subsection{Shape Moments, Moment Invariants and Shape Distributions}\label{sec:Shape_Moments}

\bigskip
\paragraph{\bf Geometric Moments.} Moments are the subject of many books
\cite{FLUSSER_Book, FLUSSER_Book2, Mundy} and review papers \cite{Nasrudin_2021,SurveyOrthogonalMoments}. Geometric moments of a density distribution function $\rho(x,y)$ in the plane are defined as the double sequence $m_{p,q}$ of Riemann integrals \cite{Rough1905}
\begin{equation}\label{geometricMoments}
m_{p, q}(\rho) := \iint  x^py^q\rho(x,y)dx dy,\quad\quad\quad p, q = 0, 1, 2, \cdots.
\end{equation}
By Weierstrass approximation theorem, the double sequence $\{m_{p, q}\}_{p, q\in\mathbb{N}}$ is \textbf{complete} in the sense that it completely characterizes piecewise continuous distribution functions $\rho(x,y)$ with bounded support, i.e. $\rho(x,y)$ is uniquely determined by its geometric moments. This is in particular the case for $\rho(x, y)$ equal to the indicatrix of a bounded domain in the plane, like the interior of a contour.  

\paragraph{\bf Central Moments.}
By setting $\bar{x} := m_{1,0}/m_{0,0}$ and $\bar{y} := m_{0, 1}/m_{0,0}$, one can defined a double sequence of translation invariant moments,  called \textbf{central moments}, as
\begin{equation}\label{central_moments}
\mu_{p, q}(\rho) := \iint (x-\bar{x})^p(y- \bar{y})^q\rho(x,y)dx dy, \quad\quad p, q = 0, 1, 2, \cdots.
\end{equation}

\paragraph{\bf Orthogonal Moments.}
The geometric moments~\eqref{geometricMoments} can be thought as projections onto a sequence of monomials, which do not form an orthogonal basis for the $L^2$-scalar product. In order to get a more efficient representation of shapes, orthogonal moments constructed as integrals over various sequences of orthogonal polynomials (for the $L^2$-scalar product) were introduced. We refer the reader to Table~1 in the review paper~\cite{Nasrudin_2021} for a list of works introducing various orthogonal moments, including Legendre moments \cite{Teague, LiaoPawlak96}, Zernike Moments \cite{Teague} and Tchebichef Moments \cite{Mukundan}, and to Fig.~2 in \cite{SurveyOrthogonalMoments} for a classification of various orthogonal moments used for image representations.

\paragraph{\bf Harmonic Moments.}\label{Harmonic}  Davis  \cite{DAVIS1977148} studied the completeness of the sequence of moments defined as
\begin{equation}\label{harmonic}
\tau_n := \iint_{D} z^n dx dy,
\end{equation}
where $D$ is a bounded region in the complex plane, and shows that a triangle is uniquely determined by $\tau_1, \tau_2, \tau_3,\tau_4$. It was shown in \cite{ReconstructingPolygons} that harmonic
moments of up to order $2n - 3$ allow to reconstruct a simply connected $n$-gon. For applications to tomography and geophysics see \cite{ReconstructingPolygons, Golub}.

\paragraph{\bf Complex moments.} Geometric moments are related to complex moments defined by \cite{DAVIS1977148}:
\begin{equation}\label{complex_moments}
c_{p, q} := \iint_{D} z^p \bar{z}^q dx dy
\end{equation}
More precisely, according to equation~(1.4) and equation~(1.5) in \cite{DAVIS1977148}
\begin{equation}
    c_{p, q} = \sum_{j = 0,k = 0}^{p, q} 
    (\ie)^{p-j}(\ie)^{q-k}\binom{p}{j}\binom{q}{k}
    m_{j+k, p+q-j-k},
\end{equation}
\begin{equation}\label{mpqcpq}
    m_{p,q} = (\ie)^n 2^{-p-q} \sum_{j = 0,k = 0}^{p, q} \binom{p}{j}\binom{q}{k}c_{j+k, p+q-j-k}.
\end{equation}

It follows that a shape can be recovered from its complex moments. Some class of shapes can be recovered from a finite number of moments \cite{Gustafsson_2000}.

\paragraph{\bf Moment Invariants for a group of transformations.}\label{sec:Invariants}
A combination of moments will be called an (absolute) \textit{Moment Invariant} for a group $G$, if it takes the same value on $\rho$ and on all its transformations $g\cdot\rho$, $g\in G$, by the action of the group $G$. The central moments constitute an example of moments invariants for the group of translations. 

The moment invariants constructed by Hu in \cite{Hu1962} for the group of rotations are widely used in shape recognition 
tasks due to their simplicity and computational efficiency \cite{CellProfiler4}. Hu also proposed a list of moment invariants for the affine group of transformations, which was corrected in \cite{Reiss}.
Moment invariants for the other groups were proposed in \cite{Hickman, Projective_Moment_Flusser, Projective_co_moments}. In order to make use of the advantages of orthogonal basis of polynomials, some authors proposed to use Zernike moments to derive moments invariants \cite{Teague, BELKASIM19911117}, and proved their completeness \cite{Wallin}. 

Abu-Mostafa and Psaltis  \cite{Abu-Mostafa_1984} used complex moments~\eqref{complex_moments} to define moment invariants for the translations and rotations, and analysed their discriminating power in practical situations.

In \cite{AbuMostafa1985ImageNB}, the role of moments in image normalization and invariant
pattern recognition was addressed.

Flusser~\cite{FLUSSER_2000} showed that the moments invariants proposed by Hu are functionally dependent and proposed a method to construct a basis of invariants for the rotation group, and proved its independence
and completeness. Flusser and Suk \cite{flusser1993pattern} designed  a complete and independent sequence of moment invariants for the  affine group of 
transformations. In \cite{SUK_FLUSSER_2011} a general method  based on representation of invariants by graphs is proposed for automatic deriving affine moment invariants and eliminating reducible and dependent invariants.

\paragraph{\bf Shape distributions.} The idea to represent the signature of an object as a shape distribution is presented in \cite{Osada} and exploited in \cite{Charon_Pierron} for convex sets. 
The characterization of a set of points by the distribution of all mutual distances has been addressed in \cite{BOUTIN}. Contrary to the case when the distances between pairs of points are known (exploited in Multidimensional Scaling technique for dimension reduction \cite{Kruskal}), there exists non-reconstructible $n$-point configurations  for any number of points and in any dimensions. However, for the vast majority of configurations, the position of points can be recovered from the distribution of mutual distances. The same holds for the distribution of areas of sub-triangles, which allows to reconstruct most configurations, though there are counterexamples.

\newpage 
\section{Proposed Approach}\label{Theory}
\subsection{Varifold Moments as unifying Framework}\label{sec:Varifold_Moments_Def}
\bigskip
\paragraph{\bf Varifold Moments.} Given a simple, closed, regular $C^2$ curve in the plane, $\gamma: [a,b] \to \mathbb{R}^2$, $t\mapsto \gamma(t)$, we introduce invariants defined as products of the following complex-valued contour integrals:
\begin{equation}
M(f;p,q,r) := \int_\gamma f(\kappa) \, z^p \overline{z}^q \left(\frac{\dot{z}}{|\dot{z}|}\right)^r |\dot{z}| \, dt,
\end{equation}
where $f$ is an arbitrary real-valued, piecewise-continuous function of the curvature $\kappa$, the powers $p, q \in \mathbb{N}$, $r\in\mathbb{Z}$, and $z = x + \ie y$. Because the integrand is homogenous of degree one in the velocities, these integrals, which we shall call {\it Varifold Moments}, are invariant under orientation-preserving reparametrizations of the curve,  hence depends only on the shape of the curve and its orientation.  Henceforth, we omit the function $f$ when it is identically equal to $1$ and write
\begin{equation}\label{Mpqr}
M_{p,q,r}(\gamma) := \int_\gamma z^p \overline{z}^q \left(\frac{\dot{z}}{|\dot{z}|}\right)^r |\dot{z}| \, dt.
\end{equation}
By equation~\eqref{cpqMpqr}, complex moments \eqref{complex_moments} can be recovered from the 
 Varifold Moments \eqref{Mpqr}. Consequently the sequence of Varifold Moments~\eqref{Mpqr} is \textbf{complete}: it characterizes  the contour uniquely, thereby making the Varifold Moments discriminating contour features.

\paragraph{\bf Mathematical Interpretation of Varifold Moments.}
A (one-dimensional) {\it varifold} in $\mathbb{R}^2$ is a continuous linear functional on the space of continuous functions on $\mathbb{R}^2 \times \mathbb{S}^1$, which we identify with integrands that are homogeneous of degree one in the velocities.

Given a one-dimensional varifold $V$ in $\mathbb{R}^2$ with compact support, we define the {\it $(p,q,r)$-moment of $V$} as
\begin{equation}
M_{p,q,r}(V) := \langle V \, | \, L_{p,q,r} \rangle,
\end{equation}
where the integrand $L_{p,q,r}$ is defined by
\begin{equation}
L_{p,q,r} := z^p \overline{z}^q \left(\frac{\dot{z}}{|\dot{z}|}\right)^r |\dot{z}|.
\end{equation} 
Basic properties of varifold moments are listed in Appendix~\ref{properties}.

\paragraph{\bf Varifold Moment Invariants.}
A key advantage of the complex notation is its behaviour under rotation. If $\gamma_\theta$ denotes the rotated curve $t \mapsto e^{i\theta}\gamma(t)$, then
\begin{equation}\label{equivariance}
\int_{\gamma_\theta} f(\kappa) \, z^p \overline{z}^q \left(\frac{\dot{z}}{|\dot{z}|}\right)^r |\dot{z}| \, dt = e^{i(p - q + r)\theta} \int_\gamma f(\kappa) \, z^p \overline{z}^q \left(\frac{\dot{z}}{|\dot{z}|}\right)^r |\dot{z}| \, dt.
\end{equation}
Consequently, any product
\begin{equation}
M(f_1;p_1,q_1,r_1) \, M(f_2;p_2,q_2,r_2) \, \cdots \, M(f_n;p_n,q_n,r_n)
\end{equation}
satisfying 
\begin{equation}\label{invariance}
    (p_1 - q_1 + r_1) + \cdots + (p_n - q_n + r_n) = 0
\end{equation} is invariant under rotations and defines a \textbf{rotation-invariant feature}.
In order to obtain features that are also invariant by translation, we translate the curve so that the centre of mass of the enclosed domain lies at the origin. If we want to add invariance by scaling,  we also normalize the area of the enclosed domain or the length of the curve (see \cite{CGT25} for the effect of different prealignment choices).

\newpage
\subsection{Geometric interpretation of discriminating Varifold Moments Invariants}\label{geometric_interpretation}
\bigskip
\paragraph{\bf Moments of Shape Distributions as VMIs.}  Although the Varifold Moments Invariants form a complete list of invariants, we are seeking a low number of discriminating features that are invariant by translation and rotation.  Some of these invariants are best described as moments of geometrically-defined statistical distributions. These one-dimensional distributions (random variables), which are themselves translation and rotation invariants of the curve, are particular in that their moments can be expressed as products of Varifold Moments. For instance, the first and second moments of the distribution  of square distances between pairs of random points in the region $D$ are given by
\begin{equation}
\frac{M_{1,2,1}(\partial D)}{M_{0,1,1}(\partial D)} \quad \textrm{and} \quad
\frac{8 M_{2,3,1}(\partial D) M_{0,1,1}(\partial D) 
+ 2 M_{1,2,1}(\partial D)^2 + 8 M_{2,1,1}(\partial D) M_{0,3,1}(\partial D)}{12 M_{0,1,1}(\partial D)^2},
\end{equation}
respectively, assuming the center of mass of the region lies at the origin. As a second example, we consider the distribution of signed areas of triangles formed by random pairs of points on the curve and the center of mass. This distribution is symmetric about zero and its odd moments vanish. Its second and
and fourth moments are given by  
$$
\frac{M_{1,1,0}(\gamma)^2 - |M_{2,0,0}(\gamma)|^2}{2M_{0,0,0}(\gamma)^2}
\quad \text{and} \quad
\frac{|M_{4,0,0}(\gamma)|^2 - 4|M_{3,1,0}(\gamma)|^2 + 3M_{2,2,0}(\gamma)^2}{8M_{0,0,0}(\gamma)^2},
$$
respectively.  The reader will find details on the computations in Appendix~\ref{App:Shape-distributions} and on properties of Varifolds Moments in Appendix~\ref{properties}. 
 Searching for discriminating invariants with a clear geometric interpretation, we found a number of shape distributions whose moments are VMIs: 
\begin{itemize}
    \item {\sl Square distance between two random points in the region.}
    \item {\sl Square distance from a random point on the region to the center of mass.}
    \item {\sl Square distance from a random point on curve to the center of mass.}
    \item {\sl Signed distance from the center of mass to a random tangent line.}
    \item {\sl Signed distance from a random point to a random tangent line.}
    \item {\sl Signed distance from the center of mass to a random normal line.}
    \item {\sl Distribution of signed areas of triangles formed by two random points in the region and the center of mass.}
\end{itemize}
The mean, variance, skewness, and kurtosis of these shape distributions are interesting geometric invariants and proved effective discriminators.

\paragraph{\bf Classical geometric invariants as VMIs.}
Some invariants such as perimeter and area are useful when size matters, while other classical invariants such as the distance between the center of mass of the curve and the center of mass of the region it encloses (\cite{Nazarov2025}) or the distance from the the cube of center of mass (considered as a complex number) to the centered third harmonic moment,
\begin{equation}
\left|\frac{\int_D (z - \textrm{center of mass)}^3 \,dxdy}{\int_D dxdy} - \left(\frac{\int_D z \,dxdy}{\int_D dxdy}\right)^3\right|,  
\end{equation}
are useful in detecting central symmetry. On the other
hand, some new invariants such as the imaginary part of 
$M_{0,2,2}(\partial D)$ or the real part of $M_{0,3,3}(\partial D)$ are
useful in detecting axial symmetry.

\paragraph{\bf Curvature-weighted Varifold Moments.}
Finally, we made use of a number of VMIs that include the curvature function and its different discretizations \cite{crane, izmestiev2025discretecurvature, Scholtes2019VariationalCO}. They include the
absolute total curvature, the Euler-Bernoulli functional (integral of the square curvature with respect to arc-length), and integrals of the form
\begin{equation}
H \! M_{p,q,r}(\gamma) := \int_\gamma \text{sgn}(\kappa) z^p \overline{z}^q \left(\frac{\dot{z}}{|\dot{z}|} \right)^r |\dot{z}| dt.
\end{equation}

Here too, the varifold interpretation is key:  a polygon can be interpreted as a varifold simply by being a collection of line segments over which to integrate, but an alternative interpretation is to integrate over the exterior angles. This viewpoint is already implicit in the definition of the total curvature of a polygon as the sum of its exterior angles and leads  to the following varifold moments:

\begin{itemize}
    \item $\sum_{j=0}^n |{\bf k}(z_{j-1},z_j,z_{j+1})|^m$,
where 
\begin{equation}
{\bf k}(z_{j-1}, z_j, z_{j+1}) := \frac{2}{|z_{j+1} - z_j| + |z_j - z_{j-1}|} \left( \frac{z_{j+1} - z_j}{|z_{j+1} - z_j|} - \frac{z_j - z_{j-1}}{|z_j - z_{j-1}|} \right),
\end{equation}
is a discretization of the mean curvature vector consisting of the difference of the two unit ``tangent" vectors at the vertex $z_j$ divided by the mean of the lengths of the edges $z_{j-1}z_j$ and $z_jz_{j+1}$,
\smallskip
\item $\sum_{j=0}^n z_j^p \overline{z}_j^p (\theta_{j+1}-\theta_j)^m \quad (m\in\mathbb{N})$,

\smallskip
\item $
\sum_{j=0}^n z_j^p \overline{z}_j^{p + r} 
\left( \left( \frac{z_{j+1} - z_j}{|z_{j+1} - z_j|}\right)^r -
\left( \frac{z_{j} - z_{j-1}}{|z_{j} - z_{j-1}|}\right)^r \right),
$
\smallskip
\item $ 
\sum_{j=0}^n \text{sgn}(\theta_{j+1} - \theta_j) \, 
z_j^p  \overline{z}_j^{p + r} 
\left( \left( \frac{z_{j+1} - z_j}{|z_{j+1} - z_j|}\right)^r -
\left( \frac{z_{j} - z_{j-1}}{|z_{j} - z_{j-1}|}\right)^r \right) .
$

\end{itemize}

 Here the polygon $\gamma$ is assumed to have vertices $z_0, \ldots,z_n$, we have set $z_n = z_{-1}$, and defined the angles recursively by setting $\theta_0 \in [0,2\pi)$ as the argument of $z_0 - z_{-1}$, and $\theta_{j+1} - \theta_j \in [0,2\pi)$ for $0 \leq j \leq n$ as the argument of $(z_{j+1} - z_j)/(z_j - z_{j-1})$.

These new varifold moments also satisfy the equivariance property~\eqref{equivariance}: when the curve is rotated by angle $\theta$, the moment is  multiplied by $e^{i(p-q+r)\theta}.$

\section{Experiments}\label{Experiments}

\subsection{Datasets}
We performed our experiments using three types of shapes: \textit{Geometric shapes and objects}, \textit{Leaves} and \textit{Cells}. We considered the following datasets 
(detailed in Appendix~\ref{datasets}): 
\begin{itemize}
    \item \textit{Datasets of Geometric shapes and objects:}
    \begin{itemize}
        \item \textbf{Mendeley:} 9 classes (triangles, circles and stars); 1000 samples per class. 
        \item \textbf{MNIST:} 10 classes (handwritten digits); 7000 samples per class.
        \item \textbf{MPEG-7:} 70 classes (animals, tools); 20 samples per class.
        \item \textbf{MPEG-400:} 20 classes (animals, household items);  20 samples per class. 
    \end{itemize}
     \item \textit{Datasets of Leaves:}
    \begin{itemize}
        \item \textbf{Swedish Leaves:} 15 classes, with 75 leaves per class.
        \item \textbf{Flavia:} 1,907 leaf images belonging to 32 classes.
    \end{itemize}
    \item \textit{Dataset of Cells}
    \begin{itemize}
        \item \textbf{BBBC010:} 2 classes with 768 individuals labeled “alive”, 639 labeled “dead”.
        \item \textbf{HeLa Kyoto:} 313 cells devided in 4 classes. 
        \item \textbf{Mouse Osteosarcoma Cells (MOC):}. 649 cells are divided in 3 classes.
    \end{itemize}
\end{itemize}

\subsection{Implementation details}\label{implementation}
Since our feature extractor is computationally lightweight (no GPU needed, see Appendix~\ref{implementation}), we extracted both scale-invariant and scale-sensitive features, yielding 63 unique VMIs as robust shape descriptors. The code for extracting the VMI features and integrating them into a classification pipeline, along with the necessary tools for contour extraction and preprocessing (e.g., arc-length parametrization), is available at the following link: \texttt{<TO ADD CODE LINK>}. In all experiments, the classifier automatically identified the most informative features without any dataset specific prior knowledge. All the details on the classifiers used (Random Forest and Multi-layer Perceptron (MPL) ) are provided in Appendix~\ref{Details on the classifiers}.

\subsection{Baselines and Evaluation}\label{Evaluation}
The proposed method is evaluated against state-of-the-art unsupervised invariant frameworks for 2D shape classification, including  ShapeEmbed \cite{romero2026shapeembed} based on ResNet-18 \cite{he2016deep}, and the recent Microsoft pipeline XShapeEncoder leveraging Zernike moments \cite{yuhheXShapeEnc2026}. Additionally, we demonstrate improvements over the classical Flusser moment descriptor \cite{flusser1993pattern} and conventional computational baselines, namely Region Property features \cite{van2014scikit} and Elliptic Fourier Descriptors \cite{persoon2007shape} (EFD). 
To evaluate the quality of the features extracted by our pipeline, we trained a Random Forest classifier \cite{RF} using a 5-fold stratified cross-validation strategy, with a 80/20 train-test split, reporting the mean and standard deviation of the $F_1$-score \cite{F1, Critic_F1} across folds. The results are reported in Tables~\ref{tab:methods_objects}, \ref{tab:methods_leaves} and \ref{tab:methods_cells}.\\

\begin{table}[ht]
\centering
\caption{Classification results using $F_1$-score (higher is better) on Objects datasets.}
\label{tab:methods_objects}
\begin{tabular}{lcccc}
\hline
Method & MNIST \cite{deng2012mnist} & MPEG-7 \cite{MPEG-7} & MPEG-400 \cite{latecki2002shape} & Mendeley \cite{KORCHI2020106090} \\
\hline
Flusser \cite{flusser1993pattern}   & $ 0.67\pm 0.00 $ & $ 0.54 \pm 0.01 $ & $  0.69    \pm 0.03 $ & $0.68 \pm 0.01 $ \\
EFD \cite{persoon2007shape} & $ 0.89 \pm 0.01 $ & $ 0.73 \pm 0.03 $ & $ 0.86 \pm 0.04 $ & $  \mathbf{1 \pm 0.00} $ \\
Region Properties \cite{van2014scikit} & $ 0.81 \pm 0.00 $ & $ 0.89 \pm 0.02 $ & $ 0.96 \pm 0.02 $ & $  0.99 \pm 0.00 $ \\
XShapeEncoder \cite{yuhheXShapeEnc2026} & $ 0.94 \pm 0.00 $ & $ 0.87 \pm 0.02 $ & $ 0.97 \pm 0.02 $ & $ \mathbf{1 \pm 0.00} $ \\
ShapeEmbed \cite{romero2026shapeembed} & $ \mathbf{0.96 \pm 0.01} $ & $0.75 \pm 0.02$ & $ 0.94 \pm 0.04  $ & $  \mathbf{1 \pm 0.00} $ \\
\textbf{VMI + RF} & $ 0.90 \pm 0.00 $ & $ \mathbf{0.93 \pm 0.02} $ & $ 0.98 \pm 0.01 $ & $ \mathbf{1 \pm 0.00} $ \\
\textbf{VMI + MLP} & $ 0.95 \pm 0.00 $ &  $ \mathbf{0.93 \pm 0.01}  $ & $ \mathbf{0.99 \pm 0.01} $ & $ 0.99 \pm 0.00 $ \\
\hline
\end{tabular}
\end{table}

\begin{table}[ht]
\centering
\caption{Classification results using $F_1$-score (higher is better) on Leaves datasets.}
\label{tab:methods_leaves}
\begin{tabular}{lcccc}
\hline
Method & Flavia \cite{wu2007leaf} & Swedish Leaves \cite{soderkvist2001swedishleaf} \\
\hline
Flusser \cite{flusser1993pattern} & $ 0.52 \pm 0.02 $ & $ 0.52 \pm 0.03 $ \\
EFD \cite{persoon2007shape}& $0.75 \pm 0.02$ & $ 0.71 \pm 0.03 $  \\
Region Properties \cite{van2014scikit}& $  0.87 \pm 0.01 $ & $ 0.84 \pm 0.02 $  \\

XShapeEncoder \cite{yuhheXShapeEnc2026} & $ 0.89 \pm 0.02 $ & $ 0.86 \pm 0.03 $  \\
ShapeEmbed \cite{romero2026shapeembed} & $0.83 \pm 0.03$ & $0.79 \pm 0.11 $  \\
\textbf{VMI + RF}  & $ \mathbf{ 0.96 \pm 0.01} $  & $\mathbf{0.96 \pm 0.01 }$  \\
\hline
\end{tabular}
\end{table}

\subsection{Sustainability versus Performance}\label{Compute}
On most of the datasets, our algorithm outperforms the other methods, with a very low computational cost (see Table~\ref{tab:times_table}).
Although ShapeEmbed give better classification rates on MNIST and BBBC010, it is computationally- and memory-intensive, and cannot be run on an ordinary laptop or embedded in a lighter devise. This limitation has at least two origins, the first one is the use of  a modified ResNet-18 \cite{he2016deep} where the standard convolutional and
pooling operations have been replaced with layers that incorporate circular padding, the second is the use of distance matrices which scale as $N^2$, where $N$ is the number of points along contours. Consequently the number of points used to discretize contours is limited  (set to $64$ in the experiments presented in \cite{romero2026shapeembed}) and an external server as to be booked to run the algorithm for many hours ($48$ hours for the MNIST dataset, whereas our algorithm runs for 3-4 hours on a computer, see Table~\ref{tab:times_table}).  We argue that the gain in performance does not justify the use of such energy demanding algorithm.

\begin{table}[ht]
\centering
\caption{Comparison of average execution time for feature extraction versus average $F_1$-score (higher is better) obtained for classification (the classification time is not counted; for VMI the classifier is MLP for MNIST,  RF for MPEG-7 and Flavia).}
\label{tab:times_table}
\begin{tabular}{lclll}
\hline
& & {MNIST} \cite{deng2012mnist}& {MPEG-7} \cite{MPEG-7} & {Flavia} \cite{wu2007leaf}\\
Method & Hardware & Duration/$F_1$  & Duration/$F_1$ & Duration/$F_1$  \\
\hline
Flusser \cite{flusser1993pattern} & CPU &  $ 32\textrm { min }  / 0.67$ &   $ 1\textrm { min } / 0.54 $&  $ 1\textrm { min } / 0.52 $  \\
EFD \cite{persoon2007shape} & CPU& $ 3\textrm { min } / 0.89 $ & $ \textbf{5\textrm { sec }} / 0.73$ & $\textbf{5\textrm { sec }} / 0.75$  \\
Region Properties \cite{van2014scikit} & CPU& $ \textbf{2\textrm { min }} / 0.81 $  & $ 2\textrm { min } / 0.89$  & $ 19\textrm { min } / 0.87$ \\
%O2V \cite{hein2025global} & $ TO \pm DO $ & $ TO \pm DO $  \\
XShapeEncoder \cite{yuhheXShapeEnc2026}  & CPU& $ 420 \textrm { min } / 0.94 $&  $ 14 \textrm{ min} / 0.87$  & $20 \textrm{ min} / 0.89$\\
ShapeEmbed \cite{romero2026shapeembed} & GPU  & $ 2,910 \textrm { min } / \textbf{0.96} $&  $ 270 \textrm { min }  / 0.75$ &  $ 270 \textrm { min } /0.83 $ \\
\textbf{VMI}  & CPU &$ 210  \textrm { min } /0.95  $ &  $ 5 \textrm{ min} / \textbf{0.93}$ &  $ 8 \textrm{ min} /\textbf{0.96}$\\
\hline
\end{tabular}
\end{table}

\subsection{Invariances Check}\label{Invariance_Check}

Since Varifold Moments are computed using cyclic sums over sample points of a curve, they are  invariant under reindexing by design. They are also stable under the change of the number of points used for the discretization of the curve (Appendix~\ref{App:discretization}), as well as under addition of Gaussian noise (Appendix~\ref{Noise}).

To evaluate the invariance of VMI under rotations and translations, we construct augmented datasets by apply random rigid transformations to three datasets, MNIST, MPEG-7 and Flavia.
Each contour has been randomly rotated by an angle $\theta_i \sim \mathcal{U}(0^\circ, 360^\circ)$ around its centroid and followed by a random translation  $(t_x^i, t_y^i) \sim \mathcal{U}(-50, 50)^2$. In Table \ref{tab:rot_invariants} we refer to them as rMNIST, rMPEG-7 and rFlavia.
From the results of Table \ref{tab:rot_invariants} we can see that our pipeline, including prealigning the center of mass at the origin, is completely invariant under rotations and translations.

\begin{table}[ht]
\centering
\caption{Classification performance measured by $F_1$-score (higher is better) on the rotated and translated datasets.}
\label{tab:rot_invariants}
\begin{tabular}{lcccc}
\hline
Method & rMNIST & rMPEG-7 & rFlavia \\
\hline
XShapeEncoder \cite{yuhheXShapeEnc2026} & $ 0.76 \pm 0.00 $ & $ 0.85 \pm 0.02 $ & $ 0.85 \pm 0.01 $ \\
ShapeEmbed \cite{romero2026shapeembed} & $0.85 \pm 0.01$ & $0.66 \pm 0.05 $ & $ 0.78 \pm 0.05 $\\
\textbf{VMI + RF}  &$\textbf{0.90} \pm 0.00 $ & $ \mathbf{0.93 \pm 0.02} $ & $ \mathbf{0.95 \pm 0.01} $ \\
\hline  
\end{tabular}
\end{table}

\subsection{Morphological Analysis of Biological Shapes}\label{Biological}
Contour analysis in biology often focuses on the study of cell shapes, a task that still requires substantial user input. Biologists typically inspect images individually, assisted by semi-automatic tools as well as manual procedures. While this approach is feasible for small datasets, it becomes impractical when dealing with millions of images, such as frames extracted from videos, where automation is essential. To address this challenge, methods such as \cite{romero2026shapeembed} and \cite{O2-VAE} have been proposed, aiming to shift shape classification from manual analysis to automated pipelines. However, these approaches do not fully alleviate the significant time required for feature extraction and accurate classification across diverse cell types. In this section, we present experiments demonstrating that our method achieves strong classification performance while reducing computational cost and processing time (Table \ref{tab:times_table}). We can observe in Table~\ref{tab:methods_cells} that VMI succeed to extract meaningful features from contours also for small and not orientable dataset, without requiring information such as texture or color. 

For the BBBC010 dataset, by exploiting a confident learning approach, we identified that approximately 7\% of the samples are incorrectly or misleadingly labeled. Examples of these mislabeled instances are shown in the Appendix, Fig.~\ref{fig:BBBC_miss}. We therefore introduce a cleaned version of the dataset, denoted cBBBC010, which is used to evaluate the different approaches; the results are reported in Table~\ref{tab:methods_cells}. We observe that the $F_1$-scores of all but one method improve on the cleaned dataset.

\begin{table}[ht]
\centering
\caption{Classification performance measured by $F_1$-score (higher is better) on Cells datasets.}
\label{tab:methods_cells}
\begin{tabular}{lcccc}
\hline
Method & BBBC010 \cite{ljosa2012annotated} &  cBBBC010 \cite{ljosa2012annotated} & HeLa Kyoto \cite{held2010cellcognition} & MOC \cite{miolane2020geomstats} \\
\hline
Flusser \cite{flusser1993pattern} & $ 0.78 \pm 0.02 $ & $ 0.85 \pm 0.02 $ & $ 0.78 \pm 0.05 $ & $ 0.67 \pm 0.02 $\\
EFD \cite{persoon2007shape} & $ 0.81 \pm 0.02 $ & $ 0.88 \pm 0.02 $ & $ 0.55 \pm 0.06 $ & $ 0.53 \pm 0.03 $ \\
Region Properties \cite{van2014scikit} & $ 0.87 \pm 0.03 $ & $ \mathbf{0.93 \pm 0.01} $ & $ 0.87 \pm 0.04 $ & $ 0.70 \pm 0.03 $  \\
XShapeEncoder \cite{yuhheXShapeEnc2026} & $ 0.83 \pm 0.02 $ & $ 0.92 \pm 0.05 $ & $ 0.72 \pm 0.04 $ & $ 0.52 \pm 0.04 $ \\
ShapeEmbed \cite{romero2026shapeembed} & $0.83 \pm 0.00$ & $ 0.84 \pm 0.03 $ & $0.80 \pm 0.04$ & $0.68 \pm 0.01$ \\
ShapeEmbed + Sz \cite{romero2026shapeembed} & $\mathbf{0.87 \pm 0.01}$ & $ 0.82 \pm 0.03 $ & $0.85 \pm 0.04$ & $0.70 \pm 0.05$ \\
\textbf{VMI + RF}  & $0.85 \pm 0.03 $ & $ 0.92 \pm 0.02 $ & $ \mathbf{0.91 \pm 0.02} $ & $\mathbf{0.83 \pm 0.02}$  \\
\hline
\end{tabular}
\end{table}

\section{Conclusion}

In this work, we introduce Varifold Moments and use them to create high discriminative invariant features for shapes representations with a clear geometric meaning. We show that they provide compact and robust representation of shapes on classifications tasks that require invariance under reparameterization, translations, rotations, and/or scalings. We test our approach on  various types of datasets (objects, leaves, cells).
The resulting contour feature extraction algorithm  coupled with light classifiers outperforms state-of-the-art algorithms with significantly lower computational costs. One limitation of our work is the handcrafted nature of the invariant features proposed. We envision that more discriminative Varifold Moment Invariants still have to be discovered. The extension of this framework to higher dimension is left  for future work.

\newpage
%\begin{ack}

%\end{ack}

\bibliographystyle{unsrt}
\bibliography{bibliography}

@inproceedings{
romero2026shapeembed,
title={Shape{E}mbed: a self-supervised learning framework for 2D contour quantification},
author={Anna Foix Romero and Craig Russell and Alexander Krull and Virginie Uhlmann},
booktitle={The Thirty-ninth Annual Conference on Neural Information Processing Systems},
year={2026},
url={https://openreview.net/forum?id=iT2ZisemFs}
}

@article{Charlier_Functional_shape,
	Author = {Benjamin Charlier and Nicolas Charon  and Alain Trouv{\'e}},
	Da = {2017/04/01},
	Doi = {10.1007/s10208-015-9288-2},
	Id = {Charlier2017},
	Isbn = {1615-3383},
	Journal = {Foundations of Computational Mathematics},
	Number = {2},
	Pages = {287--357},
	Title = {The Fshape Framework for the Variability Analysis of Functional Shapes},
	Ty = {JOUR},
	Url = {https://doi.org/10.1007/s10208-015-9288-2},
	Volume = {17},
	Year = {2017}}

@inproceedings{Bauer_Varifold_Curves,
    author = {Martin Bauer and Martins Bruveris and Nicolas Charon and Jacob Møller-Andersen},
title = {Varifold-based matching of curves via Sobolev-type Riemannian metrics},
    booktitle = {7th MICCAI workshop on Mathematical Foundations of Computational Anatomy},
    year = {2017}
}

@article{pierson2022projection,
  title={Projection-based classification of surfaces for 3D human mesh sequence retrieval},
  author={Pierson, Emery and Paiva, Juan-Carlos {\'A}lvarez and Daoudi, Mohamed},
  journal={Computers \& Graphics},
  volume={102},
  pages={45--55},
  year={2022},
  publisher={Elsevier}
}

@inproceedings{Pierson_Human_Comparison,
author = {Emery Pierson and Mohamed Daoudi and Sylvain Arguillere},
title = {3D Shape Sequence of Human Comparison and Classification Using Current and Varifolds},
year = {2022},
isbn = {978-3-031-20061-8},
publisher = {Springer-Verlag},
address = {Berlin, Heidelberg},
url = {https://doi.org/10.1007/978-3-031-20062-5_30},
doi = {10.1007/978-3-031-20062-5_30},
booktitle = {Computer Vision – ECCV 2022: 17th European Conference, Tel Aviv, Israel, October 23–27, 2022, Proceedings, Part III},
pages = {523–539},
numpages = {17},
location = {Tel Aviv, Israel}
}

@INPROCEEDINGS{Kacem_Mutual_distances,
  author={Kacem, Anis and Daoudi, Mohamed and Amor, Boulbaba Ben and Alvarez-Paiva, Juan Carlos},
  booktitle={2017 IEEE International Conference on Computer Vision (ICCV)}, 
  title={A Novel Space-Time Representation on the Positive Semidefinite Cone for Facial Expression Recognition}, 
  year={2017},
  volume={},
  number={},
  pages={3199-3208},
  doi={10.1109/ICCV.2017.345}}

@ARTICLE{Kacem_Gram_Matrices,
  author={Kacem, Anis and Daoudi, Mohamed and Amor, Boulbaba Ben and Berretti, Stefano and Alvarez-Paiva, Juan Carlos},
  journal={IEEE Transactions on Pattern Analysis and Machine Intelligence}, 
  title={A Novel Geometric Framework on Gram Matrix Trajectories for Human Behavior Understanding}, 
  year={2020},
  volume={42},
  number={1},
  pages={1-14},
  doi={10.1109/TPAMI.2018.2872564}}

@phdthesis{Pierson_thesis,
    author = {Emery Pierson},
    title = {3D and 4D Human body surface comparison and
deformation : from geometric invariants to Riemannian
shape analysis},
    school = {University of Lille, France},
    year = {2023}
}

@article{Hartman2025,
	Author = {Hartman, Emmanuel and Pierson, Emery and Bauer, Martin and Daoudi, Mohamed and Charon, Nicolas},
	Da = {2025/04/01},
	Doi = {10.1007/s11263-024-02269-3},
	Id = {Hartman2025},
	Isbn = {1573-1405},
	Journal = {International Journal of Computer Vision},
	Number = {4},
	Pages = {1999--2024},
	Title = {Basis Restricted Elastic Shape Analysis on the Space of Unregistered Surfaces},
	Ty = {JOUR},
	Url = {https://doi.org/10.1007/s11263-024-02269-3},
	Volume = {133},
	Year = {2025}}

@article{Nazarov2025,
  author = {Fedor Nazarov and Dmitry Ryabogin and Vladyslav Yaskin},
  title = {On the Maximal Distance Between the Centers of Mass of a Planar Convex Body and Its Boundary},
  journal = {Discrete Comput Geom},
  volume = {73},
  pages = {1016--1036},
  year = {2025}
}

@article{SUK_FLUSSER_2011,  
title = {Affine moment invariants generated by graph method},
journal = {Pattern Recognition},
volume = {44},
number = {9},
pages = {2047-2056},
year = {2011},
note = {Computer Analysis of Images and Patterns},
issn = {0031-3203},
doi = {https://doi.org/10.1016/j.patcog.2010.05.015},
url = {https://www.sciencedirect.com/science/article/pii/S0031320310002359},
author = {Tomá\v{s} Suk and Jan Flusser}
}

@article{flusser1993pattern,  
title = {Pattern recognition by affine moment invariants},
journal = {Pattern Recognition},
volume = {26},
number = {1},
pages = {167-174},
year = {1993},
issn = {0031-3203},
doi = {https://doi.org/10.1016/0031-3203(93)90098-H},
url = {https://www.sciencedirect.com/science/article/pii/003132039390098H},
author = {Jan Flusser and Tomá\v{s} Suk}
}

@article{SurveyOrthogonalMoments,  
author = {Qi, Shuren and Zhang, Yushu and Wang, Chao and Zhou, Jiantao and Cao, Xiaochun},
title = {A Survey of Orthogonal Moments for Image Representation: Theory, Implementation, and Evaluation},
year = {2021},
issue_date = {January 2023},
publisher = {Association for Computing Machinery},
address = {New York, NY, USA},
volume = {55},
number = {1},
issn = {0360-0300},
url = {https://doi.org/10.1145/3479428},
doi = {10.1145/3479428},
journal = {ACM Comput. Surv.},
month = nov,
articleno = {1},
numpages = {35}
}

@article{ReconstructingPolygons,  
author = {Peyman Milanfar and George C. Verghese and William Clem Karl  and  Alan S. Willsky},
title = {Reconstructing polygons from moments with connections to array processing},
year = {1995},
issue_date = {February 1995},
publisher = {IEEE Press},
volume = {43},
number = {2},
issn = {1053-587X},
url = {https://doi.org/10.1109/78.348126},
doi = {10.1109/78.348126},
journal = {Trans. Sig. Proc.},
month = feb,
pages = {432–443},
numpages = {12}
}

@article{LiaoPawlak96,   
author = {Liao, Simon X. and Pawlak, Miroslaw},
title = {On Image Analysis by Moments},
year = {1996},
issue_date = {March 1996},
publisher = {IEEE Computer Society},
address = {USA},
volume = {18},
number = {3},
issn = {0162-8828},
url = {https://doi.org/10.1109/34.485554},
doi = {10.1109/34.485554},
journal = {IEEE Trans. Pattern Anal. Mach. Intell.},
month = mar,
pages = {254–266},
numpages = {13}
}

@Article{HuMomentforCovid19,  
AUTHOR = {Moung, Ervin Gubin and Hou, Chong Joon and Sufian, Maisarah Mohd and Hijazi, Mohd Hanafi Ahmad and Dargham, Jamal Ahmad and Omatu, Sigeru},
TITLE = {Fusion of Moment Invariant Method and Deep Learning Algorithm for COVID-19 Classification},
JOURNAL = {Big Data and Cognitive Computing},
VOLUME = {5},
YEAR = {2021},
NUMBER = {4},
ARTICLE-NUMBER = {74},
URL = {https://www.mdpi.com/2504-2289/5/4/74},
ISSN = {2504-2289},
DOI = {10.3390/bdcc5040074}
}

@book{FLUSSER_Book,  
author = {Jan Flusser and Tomá\v{s} Suk and Barbara Zitová},
publisher = {John Wiley \& Sons, Ltd},
isbn = {9780470684757},
title = {Introduction to Moments},
booktitle = {Moments and Moment Invariants in Pattern Recognition},
chapter = {1-9},
pages = {1-290},
doi = {DOI:10.1002/9780470684757},
url = {https://onlinelibrary.wiley.com/doi/abs/10.1002/9780470684757.ch1},
eprint = {https://onlinelibrary.wiley.com/doi/pdf/10.1002/9780470684757.ch1},
year = {2009}
}

@book{FLUSSER_Book2,  
author = {Jan Flusser and Tomá\v{s} Suk and Barbara Zitová},
publisher = {John Wiley \& Sons, Ltd},
title = {2D and 3D Image Analysis by Moments},
year = {2016},
doi = {DOI:10.1002/9781119039402}
}

@article{Gustafsson_2000,  
doi = {10.1088/0266-5611/16/4/312},
url = {https://doi.org/10.1088/0266-5611/16/4/312},
year = {2000},
month = {aug},
publisher = {},
volume = {16},
number = {4},
pages = {1053},
author = {Björn Gustafsson and Chiyu He and Peyman Milanfar and Mihai Putinar},
title = {Reconstructing planar domains from their
moments},
journal = {Inverse Problems}
}

@article{Golub,  
author = {Golub, Gene H. and Milanfar, Peyman and Varah, James},
title = {A Stable Numerical Method for Inverting Shape from Moments},
journal = {SIAM Journal on Scientific Computing},
volume = {21},
number = {4},
pages = {1222-1243},
year = {1999},
doi = {10.1137/S1064827597328315},
URL = { 
  https://doi.org/10.1137/S1064827597328315
},
eprint = {        https://doi.org/10.1137/S1064827597328315
}
}

@article{FLUSSER_2000,  
title = {On the independence of rotation moment invariants},
journal = {Pattern Recognition},
volume = {33},
number = {9},
pages = {1405-1410},
year = {2000},
issn = {0031-3203},
doi = {https://doi.org/10.1016/S0031-3203(99)00127-2},
url = {https://www.sciencedirect.com/science/article/pii/S0031320399001272},
author = {Jan Flusser}
}

@article{DAVIS1977148,  
title = {Plane regions determined by complex moments},
journal = {Journal of Approximation Theory},
volume = {19},
number = {2},
pages = {148-153},
year = {1977},
issn = {0021-9045},
doi = {https://doi.org/10.1016/0021-9045(77)90037-5},
url = {https://www.sciencedirect.com/science/article/pii/0021904577900375},
author = {Philip J Davis}
}

@article{Hickman,
	Author = {Hickman, Mark S. },
	Da = {2012/11/01},
	Doi = {10.1007/s10851-011-0323-x},
	Id = {Hickman2012},
	Isbn = {1573-7683},
	Journal = {Journal of Mathematical Imaging and Vision},
	Number = {3},
	Pages = {223--235},
	Title = {Geometric Moments and Their Invariants},
	Ty = {JOUR},
	Url = {https://doi.org/10.1007/s10851-011-0323-x},
	Volume = {44},
	Year = {2012}}

@article{BOUTIN,  
title = {On reconstructing n-point configurations from the distribution of distances or areas},
journal = {Advances in Applied Mathematics},
volume = {32},
number = {4},
pages = {709-735},
year = {2004},
issn = {0196-8858},
doi = {https://doi.org/10.1016/S0196-8858(03)00101-5},
url = {https://www.sciencedirect.com/science/article/pii/S0196885803001015},
author = {Mireille Boutin and Gregor Kemper}
}

@ARTICLE{Integral_Invariants,  
  author={Siddharth Manay and Daniel Cremers and Byung-Woo Hong and Anthony J Yezzi Jr and Stefano Soatto},
  journal={IEEE Transactions on Pattern Analysis and Machine Intelligence}, 
  title={Integral Invariants for Shape Matching}, 
  year={2006},
  volume={28},
  number={10},
  pages={1602-1618},
  doi={10.1109/TPAMI.2006.208}}

@article{Charon_Pierron,
	Author = {Charon, Nicolas and Pierron, Thomas},
	Da = {2021/11/01},
	Doi = {10.1007/s10455-021-09795-0},
	Id = {Charon2021},
	Isbn = {1572-9060},
	Journal = {Annals of Global Analysis and Geometry},
	Number = {4},
	Pages = {863--901},
	Title = {On length measures of planar closed curves and the comparison of convex shapes},
	Ty = {JOUR},
	Url = {https://doi.org/10.1007/s10455-021-09795-0},
	Volume = {60},
	Year = {2021}}

@ARTICLE{Hu1962,  
  author={Ming-Kuei Hu},
  journal={IRE Transactions on Information Theory}, 
  title={Visual pattern recognition by moment invariants}, 
  year={1962},
  volume={8},
  number={2},
  pages={179-187},
  doi={10.1109/TIT.1962.1057692}}

@ARTICLE{Teague,  
       author = {Michael Reed Teague},
        title = {Image analysis via the general theory of moments},
      journal = {Journal of the Optical Society of America (1917-1983)},
         year = 1980,
        month = aug,
       volume = {70},
       number = {8},
        pages = {920},
          doi = {10.1364/JOSA.70.000920},
       adsurl = {https://ui.adsabs.harvard.edu/abs/1980JOSA...70..920T}
}

@article{Nasrudin_2021, 
doi = {10.1088/1742-6596/1962/1/012028},
url = {https://doi.org/10.1088/1742-6596/1962/1/012028},
year = {2021},
month = {jul},
publisher = {IOP Publishing},
volume = {1962},
number = {1},
pages = {012028},
author = {Nasrudin, Mohd Wafi and Yaakob, Nizam Shahrul and Abdul Rahim, Nazren Amir and Zahir Ahmad, Mohd Zamri and Ramli, Nuraminah and Aziz Rashid, Mohd Shaiful},
title = {Moment Invariants Technique for Image Analysis and Its Applications: A Review},
journal = {Journal of Physics: Conference Series}
}

@ARTICLE{Mukundan, 
  author={Ramakrishnan Mukundan and Seng-Huat Ong and P.A. Lee},
  journal={IEEE Transactions on Image Processing}, 
  title={Image analysis by Tchebichef moments}, 
  year={2001},
  volume={10},
  number={9},
  pages={1357-1364},
  doi={10.1109/83.941859}}

@article{Allard1972OnTF,
  title={On the first variation of a varifold},
  author={William K. Allard},
  journal={Annals of Mathematics},
  year={1972},
  volume={95},
  pages={417-491},
  url={https://api.semanticscholar.org/CorpusID:123646309}
}

@book{Santalo2004,
  title={Integral geometry and geometric probability},
  author={Santal{\'o}, Luis A},
  year={2004},
  publisher={Cambridge university press}
}

@incollection{CHARON_Fidelity,
title = {12 - Fidelity metrics between curves and surfaces: currents, varifolds, and normal cycles},
editor = {Xavier Pennec and Stefan Sommer and Tom Fletcher},
booktitle = {Riemannian Geometric Statistics in Medical Image Analysis},
publisher = {Academic Press},
pages = {441-477},
year = {2020},
isbn = {978-0-12-814725-2},
doi = {https://doi.org/10.1016/B978-0-12-814725-2.00021-2},
url = {https://www.sciencedirect.com/science/article/pii/B9780128147252000212},
author = {Nicolas Charon and Benjamin Charlier and Joan Glaunès and Pietro Gori and Pierre Roussillon}
}

@INPROCEEDINGS{Kaltenmark,
  author={Kaltenmark, Irène and Charlier, Benjamin and Charon, Nicolas},
  booktitle={2017 IEEE Conference on Computer Vision and Pattern Recognition (CVPR)}, 
  title={A General Framework for Curve and Surface Comparison and Registration with Oriented Varifolds}, 
  year={2017},
  volume={},
  number={},
  pages={4580-4589},
  doi={10.1109/CVPR.2017.487}}

@article{Mokhtarian, 
author = {Farzin Mokhtarian and Alan K Mackworth},
journal = {IEEE Transactions on Pattern Analysis and Machine Intelligence},
title = {A Theory of Multiscale, Curvature-Based Shape Representation for Planar Curves},
year = 1992,
volume = 14,
number = 8,
pages = {789--805}
}

@book{Rough1905, 
author = {Routh, Edward John},
title = {An elementary treatise on the dynamics of a system of rigid bodies},
year = {1905},
publisher = {London ; New York : Macmillan},
numpages = {564},
identifier = {ark:/13960/t9862cv9p}
}

@article{Projective_co_moments,
title = {Projective invariants of co-moments of 2D images},
journal = {Pattern Recognition},
volume = {43},
number = {10},
pages = {3233-3242},
year = {2010},
issn = {0031-3203},
doi = {https://doi.org/10.1016/j.patcog.2010.05.004},
url = {https://www.sciencedirect.com/science/article/pii/S0031320310002050},
author = {Wang Yuanbin and Zhang Bin and Yao Tianshun}
}

@ARTICLE{Projective_Moment_Flusser,
  author={Tomá\v{s} Suk and Jan Flusser},
  journal={IEEE Transactions on Pattern Analysis and Machine Intelligence}, 
  title={Projective moment invariants}, 
  year={2004},
  volume={26},
  number={10},
  pages={1364-1367},
  doi={10.1109/TPAMI.2004.89}}

@ARTICLE{Horn,
  author={Berthold K. P Horn},
  journal={Proceedings of the IEEE}, 
  title={Extended Gaussian images}, 
  year={1984},
  volume={72},
  number={12},
  pages={1671-1686},
  doi={10.1109/PROC.1984.13073}}

@article{F1,
author = {Blair, David C.},
title = {Information Retrieval, 2nd ed. C.J. Van Rijsbergen. London: Butterworths; 1979: 208 pp. Price: \$32.50},
journal = {Journal of the American Society for Information Science},
volume = {30},
number = {6},
pages = {374-375},
doi = {https://doi.org/10.1002/asi.4630300621},
url = {https://asistdl.onlinelibrary.wiley.com/doi/abs/10.1002/asi.4630300621},
eprint = {https://asistdl.onlinelibrary.wiley.com/doi/pdf/10.1002/asi.4630300621},
year = {1979}
}

@article{Osada,
author = {Osada, Robert and Funkhouser, Thomas and Chazelle, Bernard and Dobkin, David},
title = {Shape distributions},
year = {2002},
issue_date = {October 2002},
publisher = {Association for Computing Machinery},
address = {New York, NY, USA},
volume = {21},
number = {4},
issn = {0730-0301},
url = {https://doi.org/10.1145/571647.571648},
doi = {10.1145/571647.571648},
journal = {ACM Trans. Graph.},
month = oct,
pages = {807–832},
numpages = {26}
}

@article{Quach,
	Author = {Quach, Boi M. and Dinh, V. Cuong and Pham, Nhung and Huynh, Dang and Nguyen, Binh T.},
	Da = {2023/01/01},
	Doi = {10.1007/s11042-022-13199-y},
	Id = {Quach2023},
	Isbn = {1573-7721},
	Journal = {Multimedia Tools and Applications},
	Number = {1},
	Pages = {777--801},
	Title = {Leaf recognition using convolutional neural networks based features},
	Ty = {JOUR},
	Url = {https://doi.org/10.1007/s11042-022-13199-y},
	Volume = {82},
	Year = {2023}}

@article{Critic_F1,
author = {Christen, Peter and Hand, David J. and Kirielle, Nishadi},
title = {A Review of the F-Measure: Its History, Properties, Criticism, and Alternatives},
year = {2023},
issue_date = {March 2024},
publisher = {Association for Computing Machinery},
address = {New York, NY, USA},
volume = {56},
number = {3},
issn = {0360-0300},
url = {https://doi.org/10.1145/3606367},
doi = {10.1145/3606367},
journal = {ACM Comput. Surv.},
month = oct,
articleno = {73},
numpages = {24}
}

@article{RF,
	Author = {Breiman, Leo},
	Da = {2001/10/01},
	Doi = {10.1023/A:1010933404324},
	Id = {Breiman2001},
	Isbn = {1573-0565},
	Journal = {Machine Learning},
	Number = {1},
	Pages = {5--32},
	Title = {Random Forests},
	Ty = {JOUR},
	Url = {https://doi.org/10.1023/A:1010933404324},
	Volume = {45},
	Year = {2001}}

@article{Kruskal,
	Author = {Joseph B. Kruskal },
	Da = {1964/03/01},
	Doi = {10.1007/BF02289565},
	Id = {Kruskal1964},
	Isbn = {1860-0980},
	Journal = {Psychometrika},
	Number = {1},
	Pages = {1--27},
	Title = {Multidimensional scaling by optimizing goodness of fit to a nonmetric hypothesis},
	Ty = {JOUR},
	Url = {https://doi.org/10.1007/BF02289565},
	Volume = {29},
	Year = {1964}}

@book{Mundy,
author = {Joseph L. Mundy and Andrew Zisserman}, 
title = {Geometric Invariance in Computer Vision, Artifi-
cial Intelligence}, 
publisher = {MIT Press, Cambridge, MA},
year = {1992}}

@article{KORCHI2020106090,
title = {2D geometric shapes dataset – for machine learning and pattern recognition},
journal = {Data in Brief},
volume = {32},
pages = {106090},
year = {2020},
issn = {2352-3409},
doi = {https://doi.org/10.1016/j.dib.2020.106090},
url = {https://www.sciencedirect.com/science/article/pii/S2352340920309847},
author = {Anas El Korchi and Youssef Ghanou}
}

@article{deng2012mnist,
  title={The mnist database of handwritten digit images for machine learning research [best of the web]},
  author={Deng, Li},
  journal={IEEE signal processing magazine},
  volume={29},
  number={6},
  pages={141--142},
  year={2012},
  publisher={IEEE}
}

@techreport{MPEG-7,
title = {MPEG-7 Core Experiment CE-Shape-1},
institution = {ISO/IEC JTC1/SC29/WG11},
author = {Richard Ralph},
year = {1999},
url = {http://www.dabi.temple.edu/~shape/MPEG7/dataset.html},
license = {LGPL-3.0 license}
}

@mastersthesis{soderkvist2001swedishleaf,
  author       = {Oskar J. O. S{\"o}derkvist},
  title        = {Computer Vision Classification of Leaves from Swedish Trees},
  school       = {Link{\"o}ping University},
  year         = {2001},
  note         = {Swedish Leaf Dataset. Accessed: 2026-03-04},
  url          = {https://www.cvl.isy.liu.se/en/research/datasets/swedish-leaf/},
  license = {CC BY 4.0}
}

@inproceedings{wu2007leaf,
  title={A leaf recognition algorithm for plant classification using probabilistic neural network},
  author={Wu, Stephen Gang and Bao, Forrest Sheng and Xu, Eric You and Wang, Yu-Xuan and Chang, Yi-Fan and Xiang, Qiao-Liang},
  booktitle={2007 IEEE international symposium on signal processing and information technology},
  pages={11--16},
  year={2007},
  organization={IEEE},
  note = {FLAVIA dataset. Accessed: 2026-04-20},
  license = {GNU GPL}
}

@article{ljosa2012annotated,
  title={Annotated high-throughput microscopy image sets for validation},
  author={Ljosa, Vebjorn and Sokolnicki, Katherine L and Carpenter, Anne E},
  journal={Nature methods},
  volume={9},
  number={7},
  pages={637},
  year={2012},
  note = {Broad Bioimage Benchmark Collection (BBBC010)}
}

@article{held2010cellcognition,
  title={CellCognition: time-resolved phenotype annotation in high-throughput live cell imaging},
  author={Held, Michael and Schmitz, Michael HA and Fischer, Bernd and Walter, Thomas and Neumann, Beate and Olma, Michael H and Peter, Matthias and Ellenberg, Jan and Gerlich, Daniel W},
  journal={Nature methods},
  volume={7},
  number={9},
  pages={747--754},
  year={2010},
  publisher={Nature Publishing Group US New York},
  note    = {HeLa Kyoto dataset}
}

@article{miolane2020geomstats,
  title={Geomstats: A python package for riemannian geometry in machine learning},
  author={Miolane, Nina and Guigui, Nicolas and Le Brigant, Alice and Mathe, Johan and Hou, Benjamin and Thanwerdas, Yann and Heyder, Stefan and Peltre, Olivier and Koep, Niklas and Zaatiti, Hadi and others},
  journal={Journal of Machine Learning Research},
  volume={21},
  number={223},
  pages={1--9},
  year={2020},
  note    = {Mouse Osteosarcoma Cells (MOC) dataset}
}

@article{latecki2002shape,
  title={Shape similarity measure based on correspondence of visual parts},
  author={Latecki, Longin Jan and Lakamper, Rolf},
  journal={IEEE Transactions on Pattern Analysis and Machine Intelligence},
  volume={22},
  number={10},
  pages={1185--1190},
  year={2002},
  publisher={IEEE},
  note      = {MPEG-400 shape dataset}
}

@inproceedings{yuhheXShapeEnc2026,
title={Training-free Spatially Grounded Geometric Shape Encoding (Technical Report)},
author={He, Yuhang},
booktitle={arXiv:2604.07522},
year={2026}}

@ARTICLE{Abu-Mostafa_1984,
  author={Abu-Mostafa, Yaser S. and Psaltis, Demetri},
  journal={IEEE Transactions on Pattern Analysis and Machine Intelligence}, 
  title={Recognitive Aspects of Moment Invariants}, 
  year={1984},
  volume={PAMI-6},
  number={6},
  pages={698-706},
  doi={10.1109/TPAMI.1984.4767594}}

@InProceedings{Blanche_Buet_2015,
author={Buet, Blanche and Leonardi, Gian Paolo and Masnou, Simon},
editor="Aujol, Jean-Fran{\c{c}}ois
and Nikolova, Mila
and Papadakis, Nicolas",
title="Discrete Varifolds: A Unified Framework for Discrete Approximations of Surfaces and Mean Curvature",
booktitle="Scale Space and Variational Methods in Computer Vision",
year="2015",
publisher="Springer International Publishing",
address="Cham",
pages="513--524",
isbn="978-3-319-18461-6"
}

@misc{Buet2023FlagfoldsAA,
  title={Flagfolds: an approach to multi-dimensional varifolds},
  author={Blanche Buet and Xavier Pennec},
  year={2023},
  url={https://api.semanticscholar.org/CorpusID:258762398}
}

@article{varifold_approach,
	Author = {Buet, Blanche and Leonardi, Gian Paolo and Masnou, Simon},
	Da = {2017/11/01},
	Doi = {10.1007/s00205-017-1141-0},
	Id = {Buet2017},
	Isbn = {1432-0673},
	Journal = {Archive for Rational Mechanics and Analysis},
	Number = {2},
	Pages = {639--694},
	Title = {A Varifold Approach to Surface Approximation},
	Ty = {JOUR},
	Url = {https://doi.org/10.1007/s00205-017-1141-0},
	Volume = {226},
	Year = {2017}}

@article {LDDMM,
	author = {Stouffer, Kaitlin M. and Chen, Xiaoyin and Zeng, Hongkui and Charlier, Benjamin and Younes, Laurent and Trouv{\'e}, Alain and Miller, Michael I.},
	title = {xIV-LDDMM Toolkit: A Suite of Image-Varifold Based Technologies for Representing and Mapping 3D Imaging and Spatial-omics Data Simultaneously Across Scales},
	elocation-id = {2024.11.04.621983},
	year = {2024},
	doi = {10.1101/2024.11.04.621983},
	publisher = {Cold Spring Harbor Laboratory},
	URL = {https://www.biorxiv.org/content/early/2024/11/05/2024.11.04.621983},
	eprint = {https://www.biorxiv.org/content/early/2024/11/05/2024.11.04.621983.full.pdf},
	journal = {bioRxiv}
}

@article{Charon_Trouve,
author = {Charon, Nicolas and Trouv\'{e}, Alain},
title = {The Varifold Representation of Nonoriented Shapes for Diffeomorphic Registration},
journal = {SIAM Journal on Imaging Sciences},
volume = {6},
number = {4},
pages = {2547-2580},
year = {2013},
doi = {10.1137/130918885},

URL = { 
    
        https://doi.org/10.1137/130918885
    
    

},
eprint = { 
    
        https://doi.org/10.1137/130918885
}
}

@InProceedings{Partial_Matching,
author="Antonsanti, Pierre-Louis
and Glaun{\`e}s, Joan
and Benseghir, Thomas
and Jugnon, Vincent
and Kaltenmark, Ir{\`e}ne",
editor="Feragen, Aasa
and Sommer, Stefan
and Schnabel, Julia
and Nielsen, Mads",
title="Partial Matching in the Space of Varifolds",
booktitle="Information Processing in Medical Imaging",
year="2021",
publisher="Springer International Publishing",
address="Cham",
pages="123--135",
isbn="978-3-030-78191-0"
}

@article{crane,
place = {Country unknown/Code not available}, title = {A Glimpse into Discrete Differential Geometry}, url = {https://par.nsf.gov/biblio/10060772}, abstractNote = {The emerging field of discrete differential geometry (DDG) studies discrete analogues of smooth geometric objects, providing an essential link between analytical descriptions and computation. In recent years it has unearthed a rich variety of new perspectives on applied problems in computational anatomy/biology, computational mechanics, industrial design, computational architecture, and digital geometry processing at large. The basic philosophy of discrete differential geometry is that a discrete object like a polyhedron is not merely an approximation of a smooth one, but rather a differential geometric object in its own right. In contrast to traditional numerical analysis which focuses on eliminating approximation error in the limit of refinement (e.g., by taking smaller and smaller finite differences), DDG places an emphasis on the so-called “mimetic” viewpoint, where key properties of a system are preserved exactly, independent of how large or small the elements of a mesh might be. Just as algorithms for simulating mechanical systems might seek to exactly preserve physical invariants such as total energy or momentum, structure-preserving models of discrete geometry seek to exactly preserve global geometric invariants such as total curvature. More broadly, DDG focuses on the discretization of objects that do not naturally fall under the umbrella of traditional numerical analysis. This article provides an overview of some of the themes in DDG.}, journal = {Notices of the American Mathematical Society}, volume = {64}, number = {10}, author = {Crane, Keenan and Wardetzky, Max}, year = {2017}}

@article{Scholtes2019VariationalCO,
  title={Variational convergence of discrete elasticae},
  author={Sebastian Scholtes and Henrik Schumacher and Max Wardetzky},
  journal={IMA Journal of Numerical Analysis},
  year={2019},
  url={https://api.semanticscholar.org/CorpusID:119131519}
}

@misc{izmestiev2025discretecurvature,
      title={Discrete curvature}, 
      author={Ivan Izmestiev},
      year={2025},
      eprint={2502.09353},
      archivePrefix={arXiv},
      primaryClass={math.DG},
      url={https://arxiv.org/abs/2502.09353}, 
}

@article{BELKASIM19911117,
title = {Pattern recognition with moment invariants: A comparative study and new results},
journal = {Pattern Recognition},
volume = {24},
number = {12},
pages = {1117-1138},
year = {1991},
issn = {0031-3203},
doi = {https://doi.org/10.1016/0031-3203(91)90140-Z},
url = {https://www.sciencedirect.com/science/article/pii/003132039190140Z},
author = {Saad O. Belkasim and Malayappan Shridhar and Majid Ahmadi}
}

@ARTICLE{Reiss,
  author={Thomas H. Reiss},
  journal={IEEE Transactions on Pattern Analysis and Machine Intelligence}, 
  title={The revised fundamental theorem of moment invariants}, 
  year={1991},
  volume={13},
  number={8},
  pages={830-834},
  doi={10.1109/34.85675}}

@article{AbuMostafa1985ImageNB,
  title={Image Normalization by Complex Moments},
  author={Yaser S. Abu-Mostafa and Demetri Psaltis},
  journal={IEEE Transactions on Pattern Analysis and Machine Intelligence},
  year={1985},
  volume={PAMI-7},
  pages={46-55},
  url={https://api.semanticscholar.org/CorpusID:6327181}
}

@article{O2-VAE,
	Author = {Burgess, James and Nirschl, Jeffrey J. and Zanellati, Maria-Clara and Lozano, Alejandro and Cohen, Sarah and Yeung-Levy, Serena},
	Da = {2024/02/03},
	Doi = {10.1038/s41467-024-45362-4},
	Id = {Burgess2024},
	Isbn = {2041-1723},
	Journal = {Nature Communications},
	Number = {1},
	Pages = {1022},
	Title = {Orientation-invariant autoencoders learn robust representations for shape profiling of cells and organelles},
	Ty = {JOUR},
	Url = {https://doi.org/10.1038/s41467-024-45362-4},
	Volume = {15},
	Year = {2024}}

@article{persoon2007shape,
  title={Shape discrimination using Fourier descriptors},
  author={Persoon, Eric and Fu, King-Sun},
  journal={IEEE Transactions on systems, man, and cybernetics},
  volume={7},
  number={3},
  pages={170--179},
  year={2007},
  publisher={IEEE}
}

@article{van2014scikit,
  title={scikit-image: image processing in Python},
  author={Van der Walt, Stefan and Sch{\"o}nberger, Johannes L and Nunez-Iglesias, Juan and Boulogne, Fran{\c{c}}ois and Warner, Joshua D and Yager, Neil and Gouillart, Emmanuelle and Yu, Tony},
  journal={PeerJ},
  volume={2},
  pages={e453},
  year={2014},
  publisher={PeerJ Inc.}
}

@Article{CGT25,
AUTHOR = {Ciuclea, Ioana and Longari, Giorgio and Tumpach, Alice Barbora},
TITLE = {Geometric Learning of Canonical Parameterizations of 2D-Curves},
JOURNAL = {Entropy},
VOLUME = {28},
YEAR = {2026},
NUMBER = {1},
ARTICLE-NUMBER = {48},
URL = {https://www.mdpi.com/1099-4300/28/1/48},
PubMedID = {41593955},
ISSN = {1099-4300},
doi = {10.3390/e28010048}
}

@article{CellProfiler4,
	Author = {Stirling, David R. and Swain-Bowden, Madison J. and Lucas, Alice M. and Carpenter, Anne E. and Cimini, Beth A. and Goodman, Allen},
	Da = {2021/09/10},
	Doi = {10.1186/s12859-021-04344-9},
	Id = {Stirling2021},
	Isbn = {1471-2105},
	Journal = {BMC Bioinformatics},
	Number = {1},
	Pages = {433},
	Title = {CellProfiler 4: improvements in speed, utility and usability},
	Ty = {JOUR},
	Url = {https://doi.org/10.1186/s12859-021-04344-9},
	Volume = {22},
	Year = {2021}}

@article{Wallin,
author = {Wallin, \r{A}ke and K\"{u}bler, Olaf},
title = {Complete Sets of Complex Zernike Moment Invariants and the Role of the Pseudoinvariants},
year = {1995},
issue_date = {November 1995},
publisher = {IEEE Computer Society},
address = {USA},
volume = {17},
number = {11},
issn = {0162-8828},
url = {https://doi.org/10.1109/34.473239},
doi = {10.1109/34.473239},
journal = {IEEE Trans. Pattern Anal. Mach. Intell.},
month = nov,
pages = {1106–1110},
numpages = {5}
}

@InProceedings{Kumar_feature_augmentation,
author="Kumar, Dinesh
and Sharma, Dharmendra
and Goecke, Roland",
editor="Blanc-Talon, Jacques
and Delmas, Patrice
and Philips, Wilfried
and Popescu, Dan
and Scheunders, Paul",
title="Feature Map Augmentation to Improve Rotation Invariance in Convolutional Neural Networks",
booktitle="Advanced Concepts for Intelligent Vision Systems",
year="2020",
publisher="Springer International Publishing",
address="Cham",
pages="348--359",
isbn="978-3-030-40605-9"
}

@article{kuhl1982elliptic,
  title={Elliptic Fourier features of a closed contour},
  author={Kuhl, Frank P and Giardina, Charles R},
  journal={Computer graphics and image processing},
  volume={18},
  number={3},
  pages={236--258},
  year={1982},
  publisher={Elsevier}
}

@article{SVM,
	Author = {Cortes, Corinna and Vapnik, Vladimir},
	Da = {1995/09/01},
	Doi = {10.1007/BF00994018},
	Id = {Cortes1995},
	Isbn = {1573-0565},
	Journal = {Machine Learning},
	Number = {3},
	Pages = {273--297},
	Title = {Support-vector networks},
	Ty = {JOUR},
	Url = {https://doi.org/10.1007/BF00994018},
	Volume = {20},
	Year = {1995}}

@article{KNN,
author = {Evelyn Fix and JL Hodges, Jr.},
year = {1989}, 
title = {Discriminatory Analysis. Nonparametric Discrimination: Consistency Properties},
journal = {International Statistical Review / Revue Internationale de Statistique}, 
issue = {57(3)},
pages = {238--247},
url = {https://doi.org/10.2307/1403797},
}

@INPROCEEDINGS{ResNet,
  author={He, Kaiming and Zhang, Xiangyu and Ren, Shaoqing and Sun, Jian},
  booktitle={2016 IEEE Conference on Computer Vision and Pattern Recognition (CVPR)}, 
  title={Deep Residual Learning for Image Recognition}, 
  year={2016},
  volume={},
  number={},
  pages={770-778},
  doi={10.1109/CVPR.2016.90}}

@article{VGG16,
  title={Very Deep Convolutional Networks for Large-Scale Image Recognition},
  author={Karen Simonyan and Andrew Zisserman},
  journal={CoRR},
  year={2014},
  volume={abs/1409.1556},
  url={https://api.semanticscholar.org/CorpusID:14124313}
}

@ARTICLE{LeNet,
  author={Yann LeCun and Leon Bottou and Yoshua Bengio and Patrick Haffner},
  journal={Proceedings of the IEEE}, 
  title={Gradient-based learning applied to document recognition}, 
  year={1998},
  volume={86},
  number={11},
  pages={2278-2324},
  doi={10.1109/5.726791}}

@ARTICLE{Haralick,
  author={Haralick, Robert M. and Shanmugam, K. and Dinstein, Its'Hak},
  journal={IEEE Transactions on Systems, Man, and Cybernetics}, 
  title={Textural Features for Image Classification}, 
  year={1973},
  volume={SMC-3},
  number={6},
  pages={610-621},
  doi={10.1109/TSMC.1973.4309314}}

@article{Lines_E2_Equivariant_for_Radio_Galaxy,
    author = {Lines, Natalie E P and Font-Quer Roset, Joan and Scaife, Anna M M},
    title = {E(2)-equivariant features in machine learning for morphological classification of radio galaxies},
    journal = {RAS Techniques and Instruments},
    volume = {3},
    number = {1},
    pages = {347-361},
    year = {2024},
    month = {01},
    issn = {2752-8200},
    doi = {10.1093/rasti/rzae022},
    url = {https://doi.org/10.1093/rasti/rzae022},
    eprint = {https://academic.oup.com/rasti/article-pdf/3/1/347/61224963/rzae022.pdf},
}

@article{Minkowski,
	Author = {Minkowski, Hermann},
	Da = {1903/12/01},
	Doi = {10.1007/BF01445180},
	Id = {Minkowski1903},
	Isbn = {1432-1807},
	Journal = {Mathematische Annalen},
	Number = {4},
	Pages = {447--495},
	Title = {Volumen und Oberfl{\"a}che},
	Ty = {JOUR},
	Url = {https://doi.org/10.1007/BF01445180},
	Volume = {57},
	Year = {1903}}

@article{Dieleman,
    author = {Dieleman, Sander and Willett, Kyle W. and Dambre, Joni},
    title = {Rotation-invariant convolutional neural networks for galaxy morphology prediction},
    journal = {Monthly Notices of the Royal Astronomical Society},
    volume = {450},
    number = {2},
    pages = {1441-1459},
    year = {2015},
    month = {06},
    issn = {0035-8711},
    doi = {10.1093/mnras/stv632},
    url = {https://doi.org/10.1093/mnras/stv632},
    eprint = {https://academic.oup.com/mnras/article-pdf/450/2/1441/3022697/stv632.pdf},
}

@inproceedings{Bekkers_Roto_Translation,
author = {Bekkers, Erik J. and Lafarge, Maxime W. and Veta, Mitko and Eppenhof, Koen A. J. and Pluim, Josien P. W. and Duits, Remco},
title = {Roto-Translation Covariant Convolutional Networks for Medical Image Analysis},
year = {2018},
isbn = {978-3-030-00927-4},
publisher = {Springer-Verlag},
address = {Berlin, Heidelberg},
url = {https://doi.org/10.1007/978-3-030-00928-1_50},
doi = {10.1007/978-3-030-00928-1_50},
booktitle = {Medical Image Computing and Computer Assisted Intervention – MICCAI 2018: 21st International Conference, Granada, Spain, September 16-20, 2018, Proceedings, Part I},
pages = {440–448},
numpages = {9},
location = {Granada, Spain}
}

@InProceedings{Cohen_Group_Equivariant_CN,
  title = 	 {Group Equivariant Convolutional Networks},
  author = 	 {Cohen, Taco and Welling, Max},
  booktitle = 	 {Proceedings of The 33rd International Conference on Machine Learning},
  pages = 	 {2990--2999},
  year = 	 {2016},
  editor = 	 {Balcan, Maria Florina and Weinberger, Kilian Q.},
  volume = 	 {48},
  series = 	 {Proceedings of Machine Learning Research},
  address = 	 {New York, New York, USA},
  month = 	 {20--22 Jun},
  publisher =    {PMLR},
  url = 	 {https://proceedings.mlr.press/v48/cohenc16.html}
}

@article{Smets_PDE,
	Author = {Smets, Bart M. N. and Portegies, Jim and Bekkers, Erik J. and Duits, Remco},
	Da = {2023/01/01},
	Doi = {10.1007/s10851-022-01114-x},
	Id = {Smets2023},
	Isbn = {1573-7683},
	Journal = {Journal of Mathematical Imaging and Vision},
	Number = {1},
	Pages = {209--239},
	Title = {PDE-Based Group Equivariant Convolutional Neural Networks},
	Ty = {JOUR},
	Url = {https://doi.org/10.1007/s10851-022-01114-x},
	Volume = {65},
	Year = {2023}}

@InProceedings{Finzi_Lie_Group_equivariance,
  title = 	 {Generalizing Convolutional Neural Networks for Equivariance to Lie Groups on Arbitrary Continuous Data},
  author =       {Finzi, Marc and Stanton, Samuel and Izmailov, Pavel and Wilson, Andrew Gordon},
  booktitle = 	 {Proceedings of the 37th International Conference on Machine Learning},
  pages = 	 {3165--3176},
  year = 	 {2020},
  editor = 	 {III, Hal Daumé and Singh, Aarti},
  volume = 	 {119},
  series = 	 {Proceedings of Machine Learning Research},
  month = 	 {13--18 Jul},
  publisher =    {PMLR},
  url = 	 {https://proceedings.mlr.press/v119/finzi20a.html}
}

@inproceedings{Mataigne_Miolane_G_Bispectrum,
author = {Mataigne, Simon and Mathe, Johan and Sanborn, Sophia and Hillar, Christopher and Miolane, Nina},
title = {The selective G-Bispectrum and its inversion: applications to G-invariant networks},
year = {2024},
isbn = {9798331314385},
publisher = {Curran Associates Inc.},
address = {Red Hook, NY, USA},
booktitle = {Proceedings of the 38th International Conference on Neural Information Processing Systems},
articleno = {3674},
numpages = {30},
location = {Vancouver, BC, Canada},
series = {NIPS '24}
}

@article{sanborn_Miolane,
  title={A general framework for robust g-invariance in g-equivariant networks},
  author={Sanborn, Sophia and Miolane, Nina},
  journal={Advances in Neural Information Processing Systems},
  volume={36},
  pages={67103--67124},
  year={2023}
}

@article{Weiler_unireps,
author = {Maurice Weiler and Gabriele Cesa},
title = {General E(2)-Equivariant Steerable CNNs},
journal = {Advances in Neural
Information Processing Systems},
year = {2019},
pages = {14334--14345}
}

@INPROCEEDINGS{Weiler_steerable_filters,
  author={Weiler, Maurice and Hamprecht, Fred A. and Storath, Martin},
  booktitle={2018 IEEE/CVF Conference on Computer Vision and Pattern Recognition}, 
  title={Learning Steerable Filters for Rotation Equivariant CNNs}, 
  year={2018},
  volume={},
  number={},
  pages={849-858},
  doi={10.1109/CVPR.2018.00095}}

@misc{gardaa2025rotatouillerotationequivariantdeep,
      title={RotaTouille: Rotation Equivariant Deep Learning for Contours}, 
      author={Odin Hoff Gardaa and Nello Blaser},
      year={2025},
      eprint={2508.16359},
      archivePrefix={arXiv},
      primaryClass={cs.LG},
      url={https://arxiv.org/abs/2508.16359}, 
}

@article{ALgemayel_Review_ML_methods_Plants,
title = {Advances in machine learning models for plant species identification: A scoping review},
journal = {Ecological Informatics},
volume = {93},
pages = {103464},
year = {2026},
issn = {1574-9541},
doi = {https://doi.org/10.1016/j.ecoinf.2025.103464},
url = {https://www.sciencedirect.com/science/article/pii/S157495412500473X},
author = {Christina Algemayel and Dany {Abou Jaoude} and Salma Talhouk and Ibrahim Issa and Carine Ghassibe}
}

@INPROCEEDINGS{Jia_Leaf_Recognotion_Zernike_Moments,
  author={Jia, Zhuohao and Liao, Simon},
  booktitle={2023 8th International Conference on Image, Vision and Computing (ICIVC)}, 
  title={Leaf Recognition Using K-Nearest Neighbors Algorithm with Zernike Moments}, 
  year={2023},
  volume={},
  number={},
  pages={665-669},
  doi={10.1109/ICIVC58118.2023.10270642}}

@article{Anant_KNN_MomentINvariant_Texture,
author = {Anant Bhardwaj and Manpreet Kaur and Anupam Kumar},
title = {{Recognition of plants by Leaf Image using Moment Invariant and Texture Analysis}},
journal = {International Journal of Innovation and Applied Studies},
volume = {3},
year = {2013},
pages = {237--248},
issue = {1},
number = {1},
issn = {2028-9324},
url = {http://www.ijias.issr-journals.org/abstract.php?article=IJIAS-13-087-01},
abstract_html_url = {http://www.ijias.issr-journals.org/abstract.php?article=IJIAS-13-087-01},
pdf_url = {http://www.issr-journals.org/links/papers.php?journal=ijias&application=pdf&article=IJIAS-13-087-01},
document_type={Article},
source={www.issr-journals.org}
}

@INPROCEEDINGS{Lukic,
  author={Lukic, Marko and Tuba, Eva and Tuba, Milan},
  booktitle={2017 IEEE 15th International Symposium on Applied Machine Intelligence and Informatics (SAMI)}, 
  title={Leaf recognition algorithm using support vector machine with Hu moments and local binary patterns}, 
  year={2017},
  volume={},
  number={},
  pages={000485-000490},
  doi={10.1109/SAMI.2017.7880358}}

@ARTICLE{MDM,
  author={Hu, Rongxiang and Jia, Wei and Ling, Haibin and Huang, Deshuang},
  journal={IEEE Transactions on Image Processing}, 
  title={Multiscale Distance Matrix for Fast Plant Leaf Recognition}, 
  year={2012},
  volume={21},
  number={11},
  pages={4667-4672},
  doi={10.1109/TIP.2012.2207391}}

@ARTICLE{Yang_TDR,
  author={Yang, Chengzhuan and Wei, Hui},
  journal={IEEE Access}, 
  title={Plant Species Recognition Using Triangle-Distance Representation}, 
  year={2019},
  volume={7},
  number={},
  pages={178108-178120},
  doi={10.1109/ACCESS.2019.2958416}}

@article{Wang_Review,
	Author = {Wang, Zhaobin and Cui, Jing and Zhu, Ying},
	Da = {2023/05/01},
	Doi = {10.1007/s10462-022-10278-2},
	Id = {Wang2023},
	Isbn = {1573-7462},
	Journal = {Artificial Intelligence Review},
	Number = {5},
	Pages = {4217--4253},
	Title = {Review of plant leaf recognition},
	Ty = {JOUR},
	Url = {https://doi.org/10.1007/s10462-022-10278-2},
	Volume = {56},
	Year = {2023}}

@INPROCEEDINGS{CBW,
  author={Wang, Bin and Gao, Yongsheng and Sun, Changming and Blumenstein, Michael and La Salle, John},
  booktitle={2017 IEEE Conference on Computer Vision and Pattern Recognition (CVPR)}, 
  title={Can Walking and Measuring Along Chord Bunches Better Describe Leaf Shapes?}, 
  year={2017},
  volume={},
  number={},
  pages={2047-2056},
  doi={10.1109/CVPR.2017.221}}

@article{Vadori2024AutomatedCO,
  title={Automated Classification of Cell Shapes: A Comparative Evaluation of Shape Descriptors},
  author={Valentina Vadori and Antonella Peruffo and Jean-Marie Gra{\"i}c and Livio Finos and Enrico Grisan},
  journal={2025 IEEE 22nd International Symposium on Biomedical Imaging (ISBI)},
  year={2024},
  pages={1-4},
  url={https://api.semanticscholar.org/CorpusID:273798732}
}

@incollection{bisong2019logistic,
  title={Logistic regression},
  author={Bisong, Ekaba},
  booktitle={Building machine learning and deep learning models on google cloud platform: A comprehensive guide for beginners},
  pages={243--250},
  year={2019},
  publisher={Springer}
}

@inproceedings{he2016deep,
  title={Deep residual learning for image recognition},
  author={He, Kaiming and Zhang, Xiangyu and Ren, Shaoqing and Sun, Jian},
  booktitle={Proceedings of the IEEE conference on computer vision and pattern recognition},
  pages={770--778},
  year={2016}
}

@article{paszke2019pytorch,
  title={Pytorch: An imperative style, high-performance deep learning library},
  author={Paszke, Adam and Gross, Sam and Massa, Francisco and Lerer, Adam and Bradbury, James and Chanan, Gregory and Killeen, Trevor and Lin, Zeming and Gimelshein, Natalia and Antiga, Luca and others},
  journal={Advances in neural information processing systems},
  volume={32},
  year={2019}
}

@inproceedings{nair2010rectified,
  title={Rectified linear units improve restricted boltzmann machines},
  author={Nair, Vinod and Hinton, Geoffrey E},
  booktitle={Proceedings of the 27th international conference on machine learning (ICML-10)},
  pages={807--814},
  year={2010}
}

@inproceedings{ioffe2015batch,
  title={Batch normalization: Accelerating deep network training by reducing internal covariate shift},
  author={Ioffe, Sergey and Szegedy, Christian},
  booktitle={International conference on machine learning},
  pages={448--456},
  year={2015},
  organization={pmlr}
}

@article{srivastava2014dropout,
  title={Dropout: a simple way to prevent neural networks from overfitting},
  author={Srivastava, Nitish and Hinton, Geoffrey and Krizhevsky, Alex and Sutskever, Ilya and Salakhutdinov, Ruslan},
  journal={The journal of machine learning research},
  volume={15},
  number={1},
  pages={1929--1958},
  year={2014},
  publisher={JMLR. org}
}

@article{guyon2002gene,
  title={Gene selection for cancer classification using support vector machines},
  author={Guyon, Isabelle and Weston, Jason and Barnhill, Stephen and Vapnik, Vladimir},
  journal={Machine learning},
  volume={46},
  number={1},
  pages={389--422},
  year={2002},
  publisher={Springer}
}

%%%%%%%%%%%%%%%%%%%%%%%%%%%%%%%%%%%%%%%%%%%%%%%%%%%%%%%%%%%%
\newpage
\appendix

\section{Properties of Varifold Moments}\label{properties}
If $\gamma$ is a $C^1$ regular curve, the moments $M_{p,q,r}(\gamma)$ satisfy the following basic properties.

\medskip
\noindent
{\bf Property 1.} $M_{p,q,r}(\gamma)$ is independent of the parametrization of $\gamma$.

\medskip
\noindent
{\bf Property 2.} If $\gamma$ is the concatenation of curves $\gamma_0, \ldots, \gamma_n$, then
$$
M_{p,q,r}(\gamma) = \sum_{j=0}^n M_{p,q,r}(\gamma_j).
$$

\medskip
\noindent
{\bf Property 3.} The conjugate of $M_{p,q,r}(\gamma)$ is given by $M_{q,p,-r}(\gamma)$.

\medskip
\noindent
{\bf Property 4.} If $\gamma$ is a closed curve, then
$$
M_{p,0,1}(\gamma) = \int_\gamma z^p \left(\frac{\dot{z}}{|\dot{z}|}\right) |\dot{z}| \, dt = \int_\gamma z^p \, dz = 0
$$
for every $p \in \mathbb{N}$, since it is the integral of a holomorphic function over a closed path.

\medskip
\noindent
{\bf Property 5.} If $\gamma$ is a closed curve and $p$ and $q$ are natural numbers with $q > 0$, then
$$
\overline{M_{p,q,1}(\gamma)} = -\frac{q}{p+1} M_{q-1,p+1,1}(\gamma).
$$

An immediate consequence of this property is that {\sl whenever $\gamma$ is a closed curve, the varifold moment $M_{p,p+1,1}(\gamma)$ is purely imaginary.}

\medskip
\noindent
{\bf Property 6.} If $\gamma_\theta$ denotes the curve $\gamma$ rotated counter-clockwise by an angle $\theta$, then
$$
M_{p,q,r}(\gamma_\theta) = e^{i(p-q+r)\theta} M_{p,q,r}(\gamma).
$$
This follows immediately from the identity $\gamma_\theta(t) = e^{i\theta} \gamma(t)$, but has some important consequences:
\begin{itemize}
\item If for some $\theta \not\equiv 0 \pmod{2\pi}$, $\gamma_\theta = \gamma$ (i.e., the curve $\gamma$ has a non-trivial rotational symmetry) and $(p-q+r)\theta$ is not a multiple of $2\pi$, then $M_{p,q,r}(\gamma) = 0$. In particular, if the curve $\gamma$ is centrally symmetric and $p - q + r$ is odd, then $M_{p,q,r}(\gamma) = 0$.
\item If $p - q + r = 0$, then the moment $M_{p,q,r}(\gamma)$ is invariant under rotations of $\gamma$.
\item More generally, if for $1 \leq j \leq n$ the numbers $p_j$ and $q_j$ are natural numbers and the numbers $r_j$ are integers such that
$$
(p_1 - q_1 + r_1) + (p_2 - q_2 + r_2) + \cdots + (p_n - q_n + r_n) = 0,
$$
then the product $M_{p_1,q_1,r_1}(\gamma) M_{p_2,q_2,r_2}(\gamma) \cdots M_{p_n,q_n,r_n}(\gamma)$ is a rotation invariant of $\gamma$.
\end{itemize}

\medskip
\noindent
{\bf Property 7.} If $\overline{\gamma}$ denotes the image of the parametrized curve $\gamma$ under conjugation, then
$$
M_{p,q,r}(\overline{\gamma}) = \overline{M_{p,q,r}(\gamma)}.
$$

In particular, if {\it as a point set} the curve $\gamma$ is invariant under reflection across the $x$-axis, then the varifold moment $M_{p,q,r}(\gamma)$ will be real when $r$ is even and purely imaginary when $r$ is odd. 

%%%%%%%%%%%%%%%%%%%%%%%%%%%%%%%%%%%%%%%%%%%%%%%%%%%%%%%%%
\section{Geometric probabilities and varifold moments}
\label{App:Shape-distributions}

Given a region of finite area $D$ in the Euclidean plane, the statistical distribution of
mutual distances between pairs of points in $D$ defines a probability measure on the real
line that is itself a Euclidean invariant of $D$. Another example of an invariant probability
measure associated with $D$ is the statistical distribution of the areas of all triangles with
vertices in $D$. The study of these distributions is part of classical integral geometry (see, for instance, \cite{Santalo2004}) and their use in the classification of shapes was pioneered in~\cite{Osada}.

In this section we present a number of invariant statistical distributions associated with piecewise $C^1$ curves on the plane and show that their moments---which are numerical Euclidean invariants of the curve---can be computed in terms of varifold moments.

\medskip
\noindent
\textit{Distributions of mutual square distances.}
If we wish to use the distribution of
square distances between pairs of points to construct Euclidean invariants, we have the choice of taking the points in the region $D$ or restricting to pairs of points on its boundary. For both choices, their moments are easily expressed in terms of varifold moments.

Up to division by the square of the area of $D$, the $n$-th moment for the statistical distribution of square distances between pairs of
points in $D$ is given by
\begin{align*}
&\frac{-1}{4}\iint_{D \times D} |z - w|^{2n} \, d\overline{z}\wedge dz \wedge d\overline{w} \wedge dw  = \\
& \frac{-1}{4}\iint_{D \times D}
(z\overline{z} - z \overline{w} - w \overline{z} + w \overline{w})^{n} \, d\overline{z}\wedge dz \wedge d\overline{w} \wedge dw,
\end{align*}
which results in a sum of separable integrands and a sum of varifold moment invariants.
For example, momentarily forgetting the factor $-1/4$, the mean of the mutual square distances is given by
\begin{align*}
& \iint_{D \times D}
(z\overline{z} - z \overline{w} - w \overline{z} + w \overline{w})
\, d\overline{z}\wedge dz \wedge d\overline{w} \wedge dw  \\
&= \int_D z\overline{z} \, d\overline{z}\wedge dz \int_D d\overline{w}\wedge dw -
\int_D z \, d\overline{z}\wedge dz \int_D \overline{w} \, d\overline{w}\wedge dw \\
&-  \int_D \overline{z} \, d\overline{z}\wedge dz \int_D w \, d\overline{w}\wedge dw
+  \int_D w\overline{w} \, d\overline{w}\wedge dw \int_D d\overline{z}\wedge dz \\
&= M_{1,2,1}(\gamma)\, M_{0,1,1}(\gamma) - M_{1,1,1}(\gamma) \, M_{0,2,1}(\gamma).
\end{align*}
Reintroducing the factor $-1/4$ and dividing by the square of the area of $D$, we have that \textit{the mean square distance between pairs of points in $D$ is given by the expression}
$$
\frac{M_{1,2,1}(\gamma)\, M_{0,1,1}(\gamma) - M_{1,1,1}(\gamma) \, M_{0,2,1}(\gamma)}{M_{0,1,1}(\gamma)^2}.
$$
However, if the center of mass of the region lies at the origin, then
$M_{1,1,1}(\gamma) = 0$ and the expression simplifies to
$$
\frac{M_{1,2,1}(\gamma)}{M_{0,1,1}(\gamma)}.
$$
Also assuming the center of mass is at the origin, the second moment
of this distribution is given by 
$$
\frac{8 M_{2,3,1}(\gamma) M_{0,1,1}(\gamma) 
+ 2 M_{1,2,1}(\gamma)^2 + 8 M_{2,1,1}(\gamma) M_{0,3,1}(\gamma)}{12 M_{0,1,1}(\gamma)^2}.
$$
Other interesting quantities associated with the distribution of mutual square distances are its variance, skewness, and kurtosis, computed in terms of its second, third, and fourth central moments. These can all be computed in terms of varifold moments.

Up to division by the square of the perimeter of $\gamma$, the $n$-th moment for the statistical distribution of square distances between pairs of points of the curve $\gamma$ is given by
$$
\iint_{\gamma \times \gamma} |z(s_1) - w(s_2)|^{2n} \, ds_1\,ds_2.
$$
In the same way as before, this integral is a sum of separable integrands
and can be written as a sum of varifold moment invariants.
\begin{align*}
&\iint_{\gamma \times \gamma} |z(s_1) - z(s_2)|^2 \, ds_1\,ds_2 \\
&= \iint_{\gamma \times \gamma} (z(s_1)\overline{z}(s_1) - z(s_1)\overline{z}(s_2) - z(s_2)\overline{z}(s_1) + z(s_2)\overline{z}(s_2)) \,  ds_1\,ds_2 \\
&= \int_\gamma z(s_1)\overline{z}(s_1) \, ds_1
\int_\gamma ds_2 - \int_\gamma z(s_1) \, ds_1 \int_\gamma \overline{z}(s_2) \, ds_2 \\
&-
\int_\gamma \overline{z}(s_1) \, ds_1 \int_\gamma z(s_2) \, ds_2
+ \int_\gamma ds_1 \int_\gamma z(s_2)\overline{z}(s_2) \,ds_2 \\
&= 2(M_{1,1,0}(\gamma) M_{0,0,0}(\gamma) -
M_{1,0,0}(\gamma)M_{0,1,0}(\gamma)).
\end{align*}
We conclude that the mean square distance between pairs of points on
$\gamma$ is given by
$$
\frac{2(M_{1,1,0}(\gamma) M_{0,0,0}(\gamma) -
M_{1,0,0}(\gamma)M_{0,1,0}(\gamma))}{M_{0,0,0}(\gamma)^2}.
$$
The mean, variance, skewness, and kurtosis of the distribution of square distances for pairs of points on $\gamma$ proved very useful in our experiments.

Notice that \textit{the $n$-th moment of the distribution of square distances from a random point of $D$ to the origin is given by}
$$
\frac{\iint_D |z|^{2n} \, dx\,dy }{\iint_D \, dx\,dy }=
\frac{M_{n,n+1,1}(\gamma)}{(n+1)M_{0,1,1}(\gamma)}.
$$
In our experiments we have translated all curves so that their center of
mass lies at the origin, so these numbers are also Euclidean invariants.

Similarly, \textit{the $n$-th moment of the distribution of square distances from a random point on $\gamma$ to the origin is given by}
$$
\frac{\int_\gamma |z|^{2n} \, ds}{\int_\gamma ds} =
\frac{M_{n,n,0}(\gamma)}{M_{0,0,0}(\gamma)}.
$$
Centered moments, which are also Euclidean invariants, are obtained from these moments by the standard formulas.

\medskip
\noindent
\textit{Distribution of signed areas of triangles.}
Just as with mutual distances, there are various statistical distributions we can associate with areas of triangles with vertices on
a region or on its boundary.

Given three points $z_0$, $z_1$, and $z_2$ in $D$, the signed area of the
triangle $\Delta z_0z_1z_2$ is given as one-half the determinant of
the vectors $(z_1 - z_0)$ and $(z_2 - z_0)$, which can also be expressed
as $\Im((\overline{z}_1 - \overline{z}_0)(z_2 - z_0))$.

The $n$-th moment of the distribution of signed areas of triangles with
vertices in $D$ equals the integral
$$
\frac{-1}{8i}\iiint_{D \times D \times D} \Im((\overline{z}_1 - \overline{z}_0)(z_2 - z_0))^n \, d\overline{z}_1 \wedge dz_1 \wedge d\overline{z}_2 \wedge dz_2
\wedge d\overline{z}_0 \wedge dz_0
$$
divided by the product of $2^n$ and the cube of the area of $D$. Just
as in the case of mutual square distances, this integral breaks down
into a sum of integrals of separable integrands, which can be easily expressed in terms of complex or varifold moments. Since interchanging $z_1$ and $z_2$ changes the sign of the area, this distribution is
symmetric about $0$, and only even moments yield interesting invariants.
Moreover, these moments are not only invariant under the Euclidean group,
but also invariant under the group of unimodular affine transformations.

A simpler distribution of signed areas is that of triangles formed by two random points on $D$ and its center of mass, which we assume to be the origin. This distribution is also symmetric about zero, but its variance and kurtosis proved useful in our experiments.

Up to a factor of $-1/16$ and division by the square of the area of $D$, the variance
is given by the integral
\begin{align*}
&\iint_{D \times D} \Im(\overline{z}w)^2 \,
d\overline{z} \wedge dz \wedge d\overline{w} \wedge dw \\
&=
\frac{-1}{4} \iint_{D \times D} (\overline{z}w - z\overline{w})^2
d\overline{z} \wedge dz \wedge d\overline{w} \wedge dw \\
&= \frac{-1}{4}\iint_{D \times D} (\overline{z}^2w^2 - 2|z|^2|w|^2 + z^2\overline{w}^2) \,
d\overline{z} \wedge dz \wedge d\overline{w} \wedge dw \\
&= \frac{-1}{2}\int_D \overline{z}^2 \, d\overline{z} \wedge dz \int_D w^2  \, d\overline{w} \wedge dw
+ \frac{1}{2} \left(\int_D |z|^2 \, d\overline{z} \wedge dz \right)^2 \\
&= \frac{-1}{6} M_{0,3,1}(\gamma) M_{2,1,1}(\gamma) + \frac{1}{8}M_{1,2,1}(\gamma)^2.
\end{align*}
The variance of the distribution of signed areas of triangles formed by pairs
of random points in $D$ and the center of mass is given by
$$
\frac{ - M_{0,3,1}(\gamma) M_{2,1,1}(\gamma)/6 + M_{1,2,1}(\gamma)^2/8}{8 M_{0,1,1}(\gamma)^2} =
\frac{- | M_{0,3,1}(\gamma)|^2/9 + M_{1,2,1}(\gamma)^2/4}{16 M_{0,1,1}(\gamma)^2},
$$
where we used the identity $\overline{M_{0,3,1}(\gamma)} = -3M_{2,1,1}(\gamma)$.

Yet another variation on the same theme is to consider the signed areas of triangles formed by a pair of random points on the curve and the center of mass of the region, which we have taken to be at the origin. The statistical distribution of these areas is again symmetric about zero, but its variance and kurtosis proved to be effective in discriminating shapes in the experiments. The second and fourth
moments are given by
$$
-\frac{|M_{2,0,0}(\gamma)|^2 - M_{1,1,0}(\gamma)^2}{2M_{0,0,0}(\gamma)^2}
\text{ and }
\frac{|M_{4,0,0}(\gamma)|^2 - 4|M_{3,1,0}(\gamma)|^2 + 3M_{2,2,0}(\gamma)^2}{8M_{0,0,0}(\gamma)^2},
$$
respectively.

While it is interesting that the moments of these shape distributions can be computed as contour integrals along the curve, they still belong to the classical picture, requiring only classical complex moments (be they of the region or of its boundary) for their effective computation.
With varifold moments we can do more.

\medskip
\noindent
\textit{Distributions of signed distances from lines to points.}

The $n$-th moment of the distribution of signed distances from tangent lines to the origin (center of mass) is given by the integral
$$
\int_\gamma \Im\left(\frac{\overline{z}\dot{z}}{|\dot{z}|}\right)^n \, ds
$$
divided by the perimeter of the curve. It turns out that the first moment
equals twice the area of the region divided by its perimeter. Indeed,
\begin{align*}
\int_\gamma \Im\left(\frac{\overline{z}\dot{z}}{|\dot{z}|}\right)
|\dot{z}| \, dt
&= \frac{1}{2i}\int_\gamma (\overline{z} \dot{z} - z \overline{\dot{z}}) \, dt
=  \frac{1}{2i}\int_\gamma \overline{z} \, dz - \frac{1}{2i}\int_\gamma z \, d\overline{z} \\
&=
\frac{1}{i}\iint_D d\overline{z}\wedge dz,
\end{align*}
which is twice the area of the region enclosed by $\gamma$. The second moment of this distribution is given by the more interesting expression
$$
\frac{M_{1,1,0}(\gamma) - \Re(M_{0,2,2}(\gamma))}{M_{0,0,0}(\gamma)}.
$$

The $n$-th moment of the distribution of signed distances from normal lines to the origin (centre of mass) is given by the integral
$$
\int_\gamma \Re\left(\frac{\overline{z}\dot{z}}{|\dot{z}|}\right)^n \, ds
$$
divided by the perimeter of the curve. In this case, the first moment (the mean) is equal to zero. The second, third, and fourth moments
are given by rather simple formulas:
\begin{align*}
m_2 &= \frac{\Re(M_{0,2,2}(\gamma)) + M_{1,1,0}(\gamma)}{2 M_{0,0,0}(\gamma)}, \quad
m_3 = \frac{\Re(M_{0,3,3}(\gamma))}{4 M_{0,0,0}(\gamma)}, \\
m_4 &=
\frac{\Re(M_{0,4,4}(\gamma)) + 4\Re(M_{1,3,2}(\gamma)) + 3M_{2,2,0}(\gamma)}{8M_{0,0,0}(\gamma)}.
\end{align*}

One last statistical distribution we wish to present is the distribution of oriented distances between a random tangent line and a random point on the curve. The $n$-th moment of
this distribution is given by the integral
$$
\iint_{\gamma \times \gamma}
\Im\left( (\overline{z}(s_1) - \overline{z}(s_2)) \,
\frac{\dot{z}(s_1)}{|\dot{z}(s_1)|} \right)^n \, ds_1\,ds_2
$$
divided by the square of the perimeter. The mean or first moment of this distribution is again twice the area of the
region divided by its perimeter, but the
second moment is more interesting:
\begin{align*}
m_2 &= \frac{\Re\left( - M_{0,2,0}(\gamma) M_{0,0,2}(\gamma) + 2M_{0,1,2}(\gamma) M_{0,1,0}(\gamma)\right) - M_{0,0,0}(\gamma) \Re(M_{0,2,2}(\gamma))}{2M_{0,0,0}(\gamma)^2} \\
&+ \frac{M_{0,0,0}(\gamma) M_{1,1,0}(\gamma) - |M_{1,0,0}(\gamma)|^2}{M_{0,0,0}(\gamma)^2}.
\end{align*}

\section{Implementation details}

\subsection{Computer resources}\label{implementation_details}
This work was implemented in Python 3.10.12 and is publicly available at <link to the code>. Unless otherwise specified, all experiments were conducted on a machine equipped with a 12th Gen Intel Core i5-12500 CPU and no GPU. Experiments involving ShapeEmbed \cite{romero2026shapeembed}, a variational autoencoder-based method, were performed on our institutional computing cluster, using an NVIDIA A40 GPU with 46 GB of GDDR6 memory (CUDA 13.2) and an AMD EPYC 7413 24-Core processor. 

\subsection{Details on the classifiers}\label{Details on the classifiers}
The first classifier used is a Random Forest (\textbf{RF}) \cite{RF} with $800$ trees, grown without depth restriction.  
At each node, the number of features considered for splitting is set to $\sqrt{p}$, where $p$ is the total number of features. The minimum number of samples required to split an internal node and to form a leaf are both set to their default values of $2$ and $1$ respectively. Bootstrap sampling is enabled, meaning that each tree is trained on a random subset of the training data drawn with replacement. Since the same features, when computed with area normalization, can be highly correlated with the original ones, we remove those whose pairwise correlation exceeds a fixed threshold. Finally, we apply Recursive Feature Elimination \cite{guyon2002gene} to select the most informative invariants for the classification task.
%\medskip
The second classifier is a Multi-Layer Perceptron (MLP) implemented in PyTorch \cite{paszke2019pytorch}. The network consists of two hidden layers followed by an output layer. The first hidden layer maps the input of dimension (number of invariants considered) to $256$ neurons, 
and the second hidden layer reduces the representation to $128$ neurons and the third one to $64$, before the final output layer produces a score for each of the number of classes to classify. Each hidden layer follows the same structure: a fully connected linear transformation, Batch Normalization \cite{ioffe2015batch} to stabilize and accelerate training, a ReLU activation function \cite{nair2010rectified} to introduce non-linearity, and a Dropout layer \cite{srivastava2014dropout} with rate $p = 0.4$ to reduce overfitting. The output layer is a plain linear projection with no activation, as the loss function is expected to handle the final normalization. 
The network is trained by minimizing the Cross-Entropy loss for several hundred epochs, with early stopping based on validation accuracy and a fixed patience parameter, the best-performing model is retained. Given the relatively small size of the network, it can typically be trained for several epochs without requiring a GPU.

\newpage
\section{Additional Detail on Datasets}\label{datasets}

\subsection{Datasets of Geometric Shapes and Objects}
\bigskip

\paragraph{\bf Mendeley} The Mendeley 2D shape benchmark \cite{KORCHI2020106090} contains 9 classes of geometric objects (Triangle, Square, Pentagon, Hexagon, Heptagon, Octagon, Nangon, Circle and Star), each class instantiated with 10,000 samples. We examined a subset of the original dataset consisting of 1,000 randomly selected shapes per class (the selected seed will be available with the rest of the code) that exhibited significant variations in scaling, rotation, and deformation. This subset offers sufficient variability in terms of transformations while reducing the computational effort required for the experiments. \\

\paragraph{\bf MNIST} The MNIST dataset (\cite{deng2012mnist}) released under the GNU GPL, contains 70,000 handwritten digit from 0 to 9 in grayscale images with roughly 7,000 samples for each class. \\

\paragraph{\bf MPEG-7} The MPEG-7 CE-Shape-1 Part B dataset~\cite{MPEG-7}, distributed under the LGPL-3.0 license, contains 1,400 binary masks of objects (animals, tools, and symbolic shapes, ect...) organized into 70 classes, with 20 samples in each class. \\

\paragraph{\bf MPEG-400} The MPEG-400 dataset~\cite{latecki2002shape} contains 400 binary silhouettes (animals, tools, household items)  grouped into 20 classes, each containing 20 samples. Notable variations in articulation, deformation, and viewing angle, make this dataset challenging for contour-based algorithms.

\subsection{Datasets of Leaves:}
\bigskip

\paragraph{\bf Swedish Leaves} The Swedish leaves dataset \cite{soderkvist2001swedishleaf} from the Linköpling University, freely downloadable at \hyperlink{https://www.cvl.isy.liu.se/en/research/datasets/swedish-leaf/}{https://www.cvl.isy.liu.se/en/research/datasets/swedish-leaf/} under license CC BY 4.0,
consists in pictures of leaves organized into 15 classes, with 75 leaves per class. This dataset is particularly challenging since  leaves with peduncle and leaves without peduncle are present in the same class \cite{CGT25}.\\ 

\paragraph{\bf Flavia}. The Flavia dataset \cite{wu2007leaf} contains 1,907 leaf images belonging to
32 classes,released under the GNU GPL and is available at \hyperlink{https://flavia.sourceforge.net/}{https://flavia.sourceforge.net/}. Achieving high classification accuracy on this dataset is challenging due to the larger number of classes and the extremely similar shapes among many of them. This dataset is particularly interesting after the samples have been randomly rotated, as it exhibits an orientation bias; that is, samples from the same class tend to be oriented in the same direction \cite{CGT25}.

\subsection{Datasets of Cells}

\paragraph{\bf BBBC010}. The Broad Bioimage Benchmark Collection 10 \cite{ljosa2012annotated} (BBBC010, no license) is a biological imaging dataset. 
%designed to test phenotypic profiling at the whole-organism level. 
It contains  1,407 individual binary masks of C. elegans nematodes divided into two classes, with   768 individuals labeled ``alive'' and 639 individuals labeled ``dead''.\\

\paragraph{\bf HeLa Kyoto} The HeLa Kyoto dataset \cite{held2010cellcognition} (CellCognition project, CC-BY 4.0 License) consists of fluorescence microscopy images of H2B-mCherry-stained HeLa Kyoto cell nuclei.

Already segmented and cropped masks for individual nuclei are available under the “classifier data” section of the dataset. We consider 313 objects in 4 classes that are representative of nuclei at key phases of mitosis: early anaphase (earlyana, 40 samples), lateana (lateana, 83 samples), metaphase (meta, 110 samples), and prometaphase (prometa, 80 samples).\\

\paragraph{\bf Mouse Osteosarcoma Cells (MOC)} The MOC dataset (MIT license, \cite{miolane2020geomstats}) consists of fluorescence microscopy images of mouse osteosarcoma cells. In this dataset, cells have been exposed to cytoskeletal perturbation through treatment with the single drugs jasplakinolide (Jasp) and cytochalasin D (Cytd). The 649 cells are divided in 3 classes: a control class with 318 untreated cells, a Cytd class with 175 cells, and a Jasp class with 156 cells.\\

\subsection{Sample images}
\begin{figure}[ht]
     \centering
     \includegraphics[width=\linewidth]{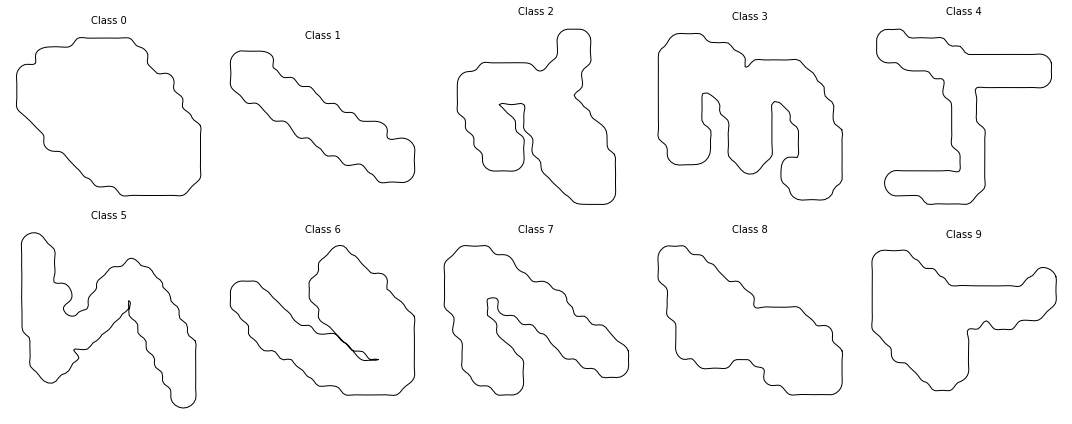}
     \caption{Visualization of random selected MNIST \cite{deng2012mnist} shapes, one per class.  }
     \label{fig:MNIST_vis}
 \end{figure}

\begin{figure}[h]
    \centering
    \includegraphics[width=\linewidth]{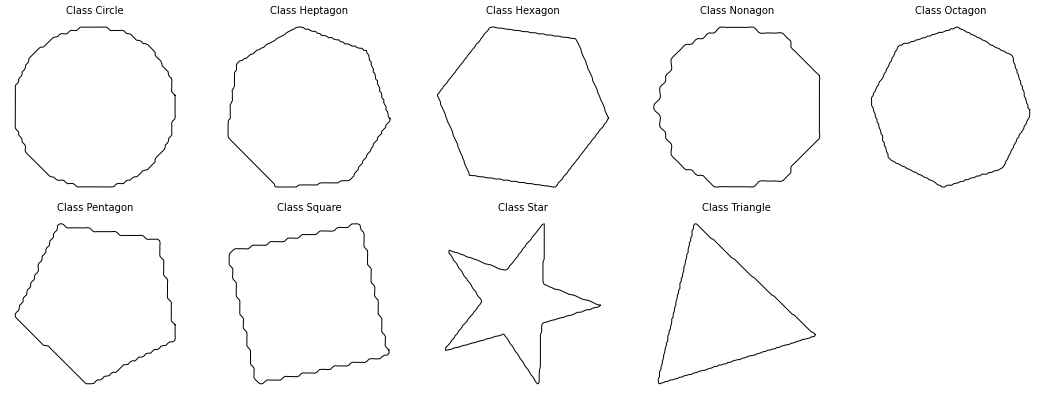}
    \caption{Visualization of random selected Mendeley \cite{KORCHI2020106090} shapes, one per class.  }
    \label{fig:mendeley_vis}
\end{figure}

\begin{figure}[H]
    \centering
    \includegraphics[width=0.45\linewidth]{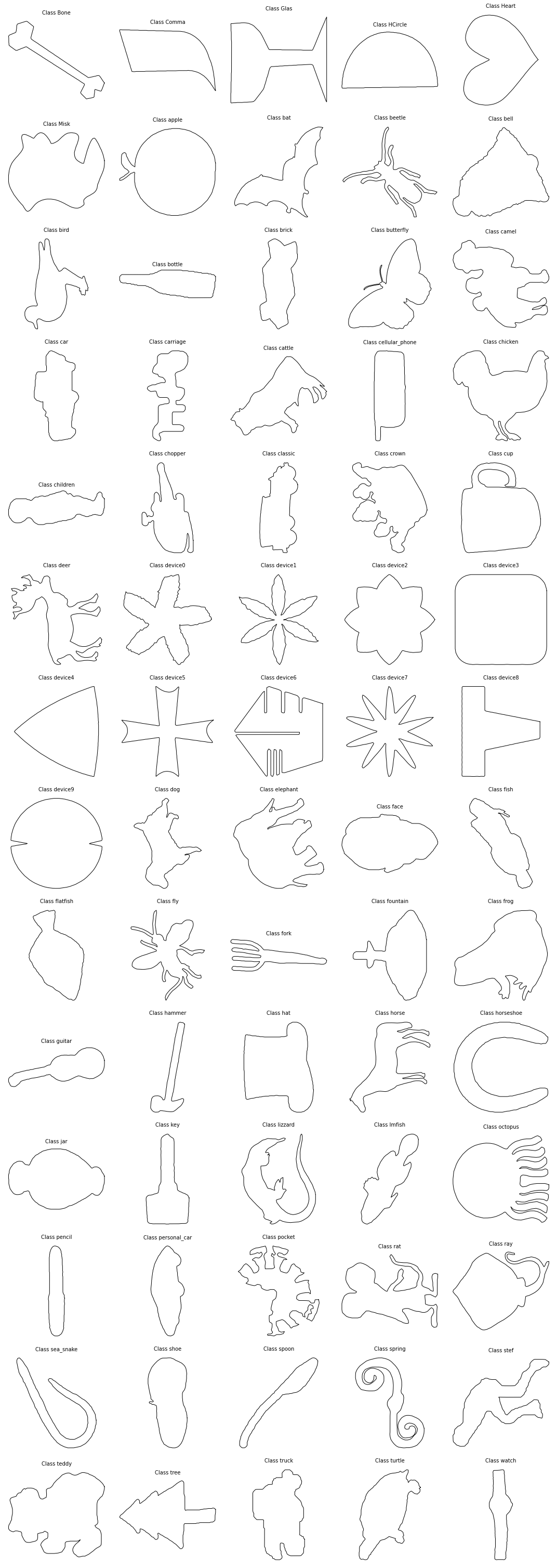}
    \caption{Visualization of random selected MPEG7 \cite{MPEG-7} shapes, one per class.  }
    \label{fig:MPEG7_vis}
\end{figure}

\begin{figure}[H]
    \centering
    \includegraphics[width=\linewidth]{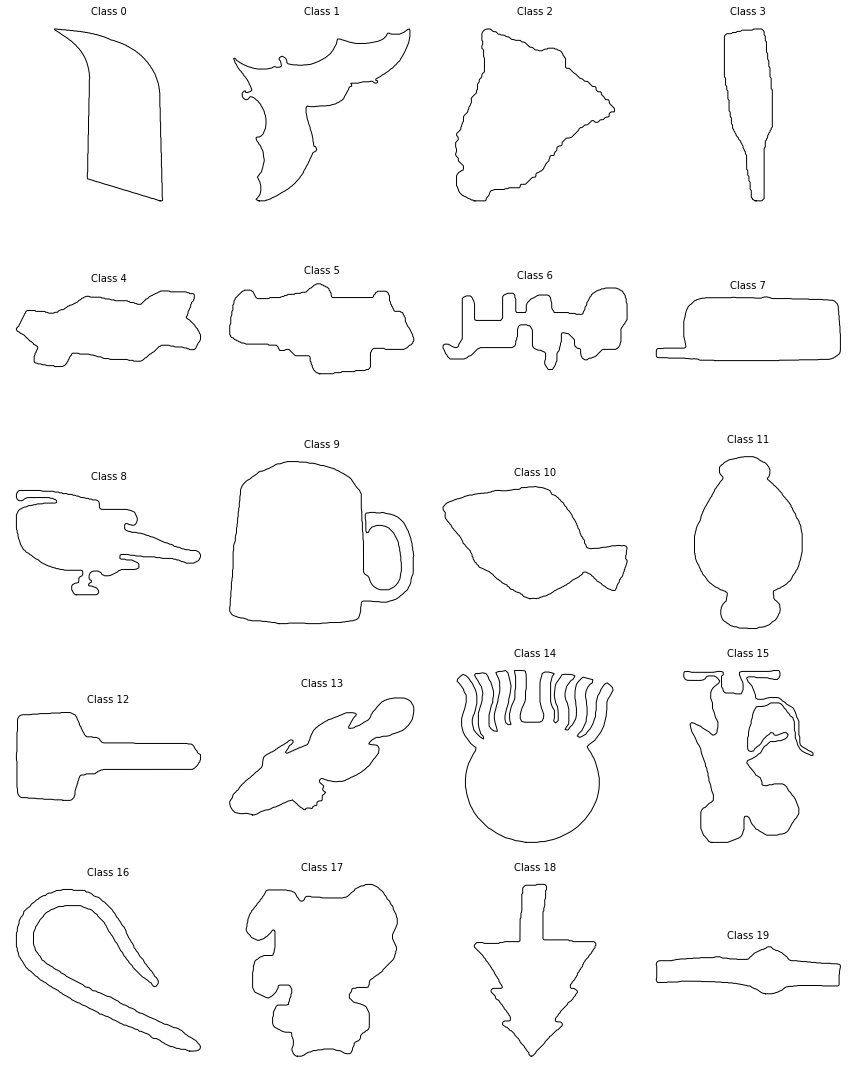}
    \caption{Visualization of random selected MPEG400 \cite{latecki2002shape} shapes, one per class.  }
    \label{fig:MPEG400_vis}
\end{figure}

\begin{figure}[H]
    \centering
    \includegraphics[width=0.6\linewidth]{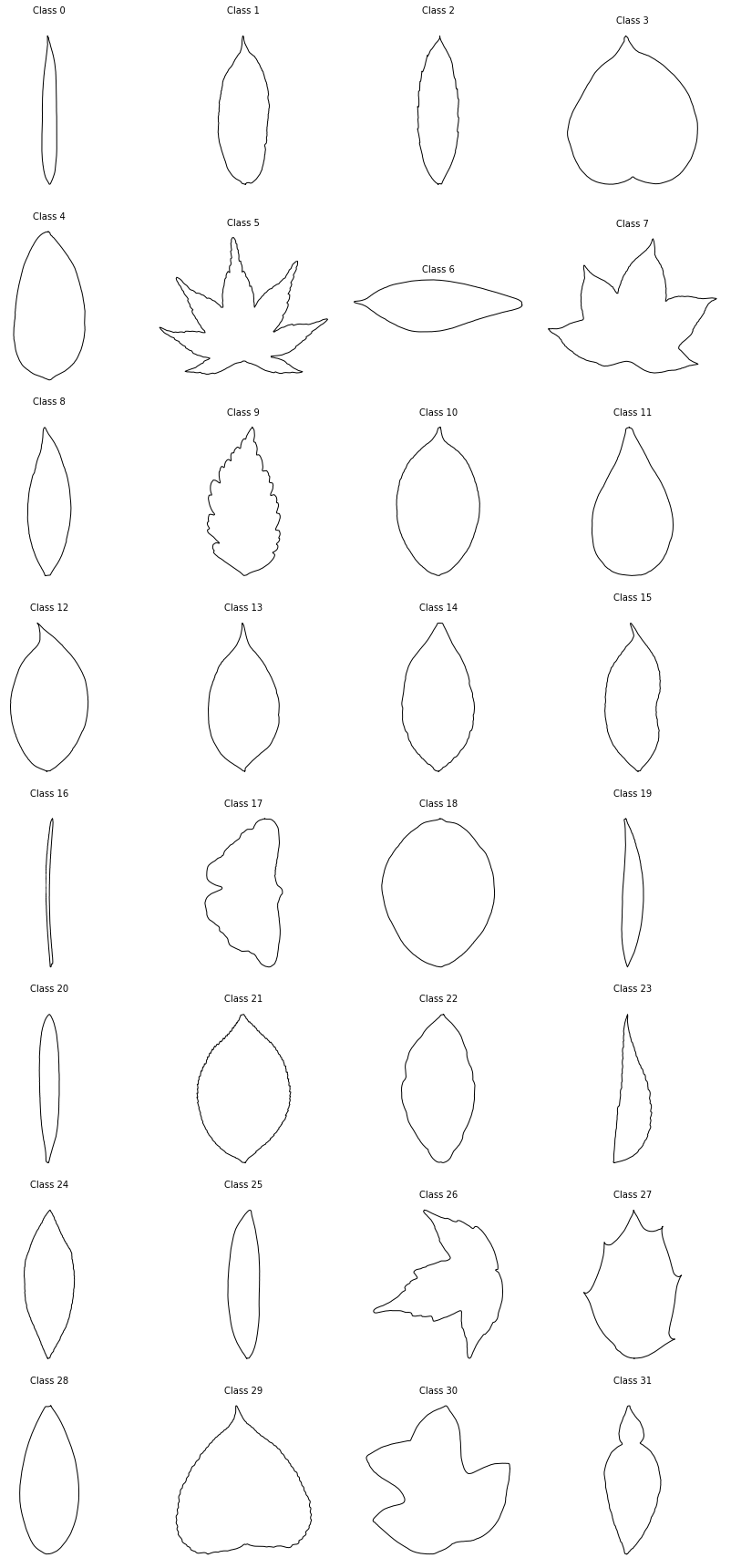}
    \caption{Visualization of random selected Flavia \cite{wu2007leaf} shapes, one per class.  }
    \label{fig:flavia_vis}
\end{figure}

\begin{figure}[H]
    \centering
    \includegraphics[width=0.9\linewidth]{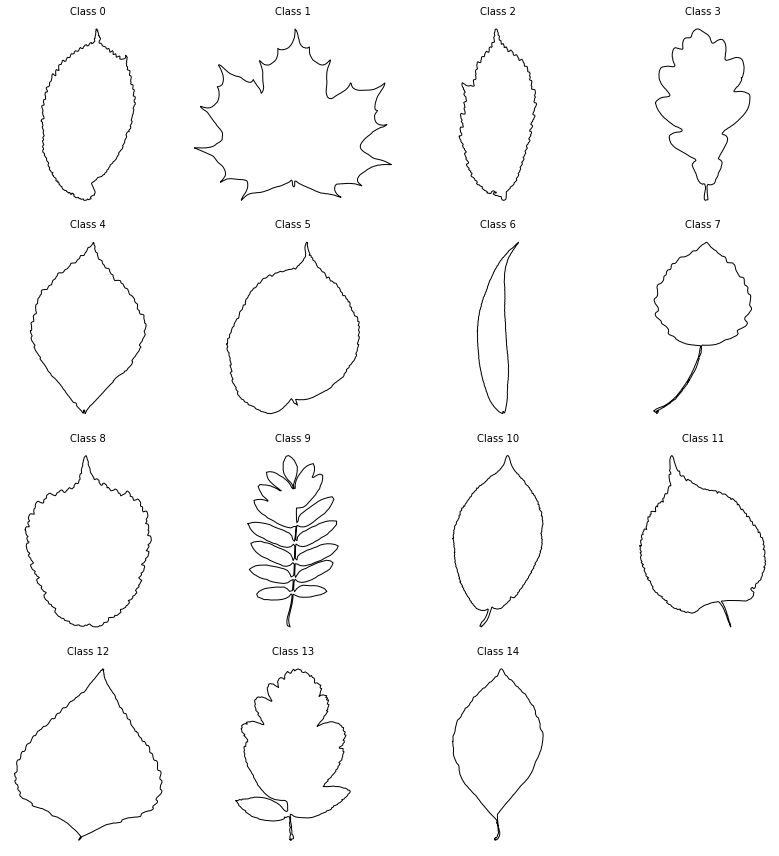}
    \caption{Visualization of random selected Swedish Leaves \cite{soderkvist2001swedishleaf} shapes, one per class.  }
    \label{fig:swedish_vis}
\end{figure}

\begin{figure}[H]
    \centering
    \includegraphics[width=0.4\linewidth]{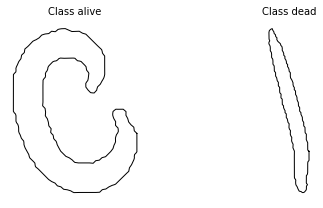}
    \caption{Visualization of random selected BBBC010 \cite{ljosa2012annotated} shapes, one per class.  }
    \label{fig:BBBC_vis}
\end{figure}

\begin{figure}[H]
    \centering
    \includegraphics[width=0.8\linewidth]{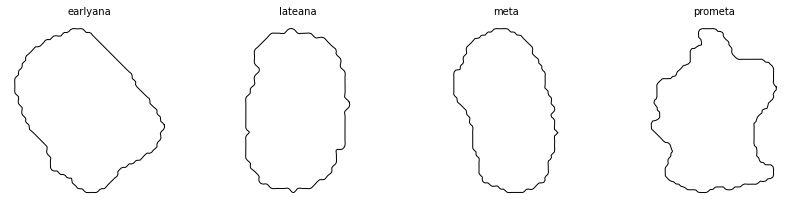}
    \caption{Visualization of random selected HeLa Kyoto \cite{held2010cellcognition} shapes, one per class.  }
    \label{fig:HK_vis}
\end{figure}

\begin{figure}[H]
    \centering
    \includegraphics[width=0.8\linewidth]{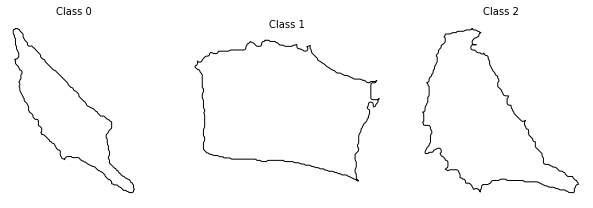}
    \caption{Visualization of random selected MOC \cite{miolane2020geomstats} shapes, one per class.  }
    \label{fig:MOC_vis}
\end{figure}

\begin{figure}[H]
    \centering
    \includegraphics[width=0.95\linewidth]{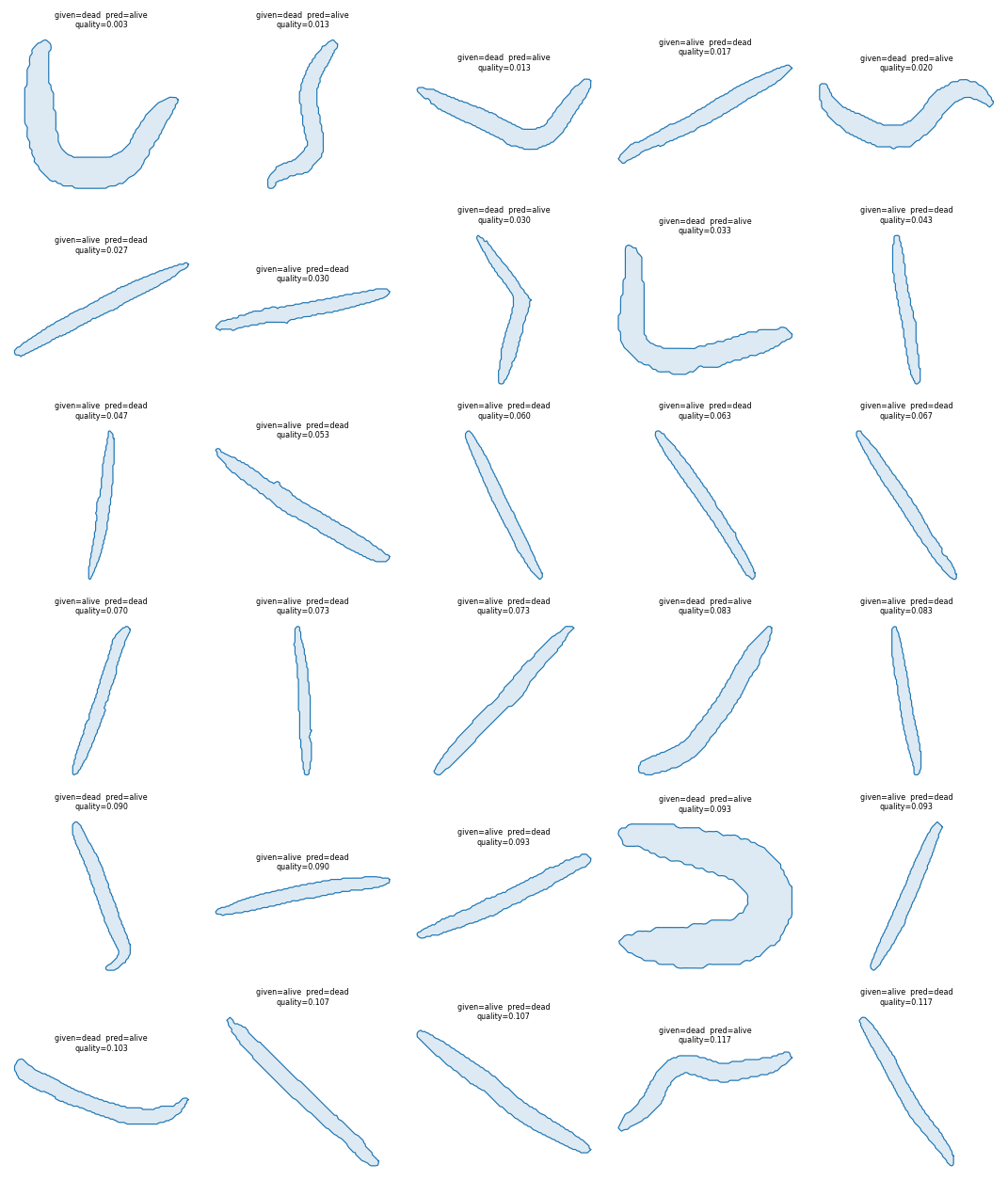}
    \caption{BBBC010 \cite{ljosa2012annotated} samples identified as potentially mislabeled using a confident learning approach. For each sample, we report the assigned label (given), the model prediction (pred), and the associated confidence score (quality). }
    \label{fig:BBBC_miss}
\end{figure}

We provide some images of the different shapes considered inside each dataset in Fig.~\ref{fig:MNIST_vis}, Fig.~\ref{fig:MPEG7_vis}, Fig.~\ref{fig:MPEG400_vis}, Fig.~\ref{fig:mendeley_vis},  Fig.~\ref{fig:flavia_vis}, Fig.~\ref{fig:swedish_vis}, Fig~\ref{fig:BBBC_vis}, Fig~\ref{fig:HK_vis}, Fig.~\ref{fig:MOC_vis}. These were extracted with different pipelines. Some dataset provided already the contour shapes (MOC \cite{miolane2020geomstats}), other the 2D masks (MPEG-7 \cite{MPEG-7}, MPEG-400 \cite{latecki2002shape}, BBBC010 \cite{ljosa2012annotated} and HeLa Kyoto\cite{held2010cellcognition}) and from them we utilized the ShapeEmbed \cite{romero2026shapeembed} pipeline to extract the boundaries, in order to have the most fair comparison. For the leaves datasets, we extracted directly the 2D contours from the images, relying on our pipeline, that will be shared with the rest of the code. In particular, for the BBBC010 dataset, by exploiting a confident learning approach, we identified that approximately 7\% of the samples are incorrectly or misleadingly labeled. Examples of these mislabeled instances are shown in Fig.~\ref{fig:BBBC_miss}. We therefore introduce a cleaned version of the dataset, denoted cBBBC010, which is used to evaluate the different approaches; the results are reported in Table~\ref{tab:methods_cells}.

\newpage

\newpage 

\section{Baselines Implementation Details}\label{sec:Baselines Implementation Details}
%\subsection{Moments Invariants}
\paragraph{\bf Flusser}
To provide a direct comparison with classical moment-based approaches, we implement the set of affine moment invariants introduced by Flusser~\cite{flusser1993pattern}. These invariants are computed using the same complex-moment framework adopted for our Variance Moment Invariants (VMI), ensuring a consistent computational pipeline across methods. In our notation, the complex moment of order $(p,q)$  is denoted by $\tfrac{1}{2i(q+1)} M_{p,q+1,1}$. The eleven Flusser invariants are then defined as follows:
\begin{itemize}
    \item $F_1 = \frac{M_{1,2,1}}{4\ie}$
    \item $F_2 = \frac{M_{2,2,1}}{{4i}} \cdot \tfrac{M_{1,3,1}}{6\ie}$
    \item $F_3 = \Re\left(\frac{M_{2,1,1}}{{2\ie}}\cdot (\frac{M_{1,3,1}}{6\ie})^2)\right)$
    \item $F_4 = \Im\left(\frac{M_{2,1,1}}{{2\ie}} \cdot (\frac{M_{1,3,1}}{6\ie})^2\right)$
    \item $F_5 = \Re\left(\frac{M_{3,1,1}}{{2\ie}} \cdot (\frac{M_{1,3,1}}{6\ie})^3\right)$
    \item $F_6 = \Im\left(\frac{M_{3,1,1}}{{2\ie}} \cdot (\frac{M_{1,3,1}}{6\ie})^3\right)$
    \item $F_7 = \frac{M_{2,3,1}}{{6\ie}} $
    \item $F_8 = \Re\left(\frac{M_{3,2,1}}{{4\ie}} \cdot (\frac{M_{1,3,1}}{6\ie})^2\right)$
    \item $F_9 = \Im\left(\frac{M_{3,2,1}}{{4\ie}} \cdot (\frac{M_{1,3,1}}{6\ie})^2\right)$
    \item $F_{10} = \Re\left(\frac{M_{4,1,1}}{{2\ie}} \cdot (\frac{M_{1,3,1}}{6\ie})^4\right)$
    \item $F_{11} = \Re\left(\frac{M_{4,1,1}}{{2\ie}} \cdot (\frac{M_{1,3,1}}{6\ie})^4\right)$
\end{itemize}

\paragraph{\bf EFD}

The Elliptical Fourier Descriptors \cite{persoon2007shape} were calculate utilizing the python library \textsc{PyEFD} (\url{https://pyefd.readthedocs.io/en/latest/}).
Specifically, the contour coordinates are approximated by a truncated Fourier series, retaining the first $K$ harmonics, where $K=10$ in all experiments. Each harmonic contributes four coefficients, yielding a feature vector of dimension 4K=40. To ensure invariance to translation, rotation, and scale, the descriptors are normalized following the standard EFD normalization procedure.
\paragraph{Region Properties}

Region based shape descriptors are used as a classical baseline by extracting a set of geometric and moment based features from each segmented object. Given a closed contour, we first rasterize it into a binary mask. 
From each mask, we compute a collection of 19 standard morphological descriptors using the \textsc{regionprops} implementation from \textsc{scikit-image} \cite{van2014scikit}. These features capture complementary aspects of object geometry, including size, elongation, convexity, and spatial extent. %To remove the next?
Specifically, the descriptor vector comprises: area, convex area, perimeter, major and minor axis lengths, extent, eccentricity, solidity, maximum Feret diameter, the seven invariant Hu moments, bounding-box height, bounding-box width, and bounding-box area. These quantities provide a compact summary of both global shape and spatial distribution.

\paragraph{\bf XShapeEncoder}

The method proposed by \cite{yuhheXShapeEnc2026} introduces both equivariant and invariant approaches for contour embedding. In our experiments, we primarily employ the equivariant network implementation provided by the author at the following link \url{https://github.com/yuhanghe01/XShapeEnc}
We further adapt the original code to extract invariant features, following the procedure described in the paper. Specifically, we retain only the magnitude of the Zernike basis projections. This modification reduces the feature dimensionality (absolute value instead of the real and imaginary components) and prevents the decoder from reconstructing the input shape in its original pose.

\paragraph{\bf ShapeEmbed} To evaluate our main competitor, ShapeEmbed \cite{romero2026shapeembed}, we aligned our experimental protocol with that described in their work and we relied on the code available at \url{https://github.com/uhlmanngroup/ShapeEmbed}. Specifically, each contour was represented using 64 sampled points. For the MOC dataset, as done in the main paper, we utilized a 512 x 512 distance matrix, and the latent space dimension fixed to 128. For each dataset, we used an 80\%/20\% train-test split and evaluated classification performance using five-fold stratified cross-validation and tested them with the provided logistic regression \cite{bisong2019logistic}.  \\

\section{Robustness to Boundary Perturbations under Gaussian Noise.}\label{Noise}
\begin{figure}[t]
    \centering
    \begin{overpic}[width=\linewidth,  ]{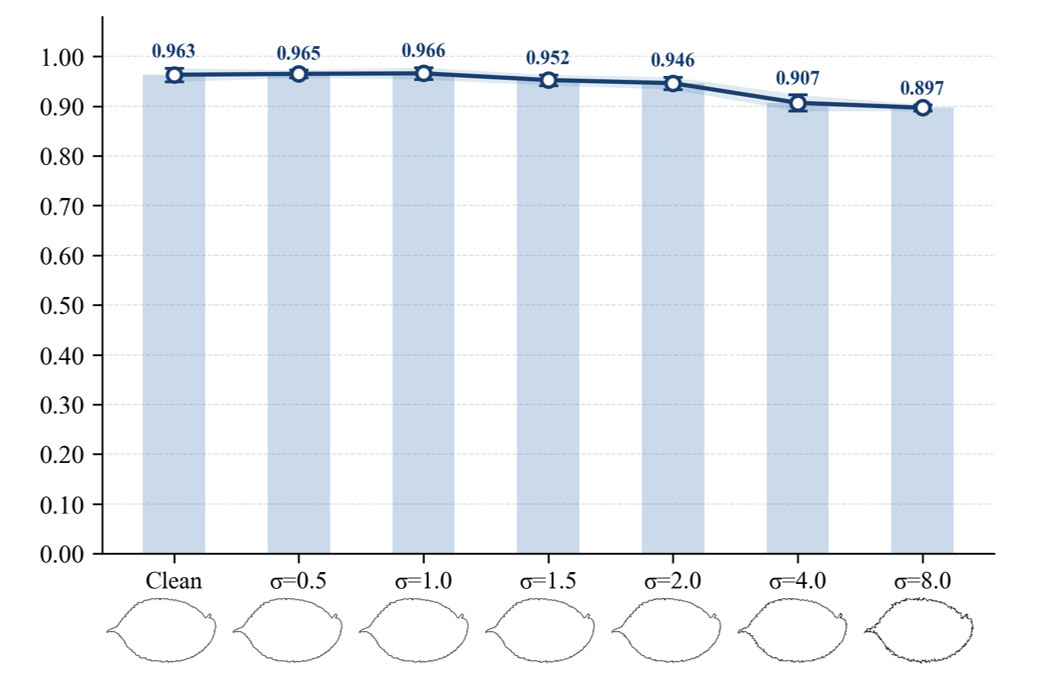} %grid, tics=10
        % X-axis label
        \put(52, 0){\makebox(0,0){\small Noise Level ($\sigma$)}}
        % Y-axis label
        \put(1.7, 40){\rotatebox{90}{\makebox(0,0){\small F1 Score}}}
    \end{overpic}
    \caption{Robustness of VMI features in a classification task on the Swedish Leaves dataset \cite{soderkvist2001swedishleaf} under Gaussian noise perturbations applied to the shape contours, evaluated in terms of $F_1$-score (Higer the better).}
    \label{fig:noise}
\end{figure}

To evaluate the sensitivity of the proposed pipeline to shape corruption, we subject the complex contour representations to additive Gaussian 
noise of increasing magnitude. Formally, given a complex contour $\gamma \subset \mathbb{C}$, we construct perturbed variants 
$\tilde{\gamma} = \gamma + \varepsilon$, where $\varepsilon \sim \mathcal{CN}(0, \sigma^2)$ is drawn independently for each 
boundary point, with $\sigma \in \{0.5, 1.0, 1.5, 2.0, 4.0, 8.0\}$. 
As illustrated in Figure ~\ref{fig:noise}, low amplitude perturbations, up to($\sigma = 1$), introduce subtle boundary irregularities while preserving global morphology, whereas higher noise levels ($\sigma > 2 $) progressively erode class-discriminative geometric cues such as global structure, margin serrations, and overall silhouette. For each noise level, we extracted the features with VMI from the Swedish Leaves dataset and tested utilize the same Random Forest approach of all the other experiments. 
Performance is reported in terms of averaged Accuracy.

\section{Robustness under successive approximations of the contour}\label{App:discretization}

To investigate the dependence of classification performance, and consequently feature quality, on the curve parameterization, we varied the number of vertices sampled along each contour and extracted features using the proposed pipeline. Table~\ref{tab:parametrization} reports the resulting classification performance for different number of equidistributed points along the contours. Additionally, it illustrates how the computational cost of the method scales with the number of sampled  points. We observe that:
\begin{itemize}
    \item Once the contour is sampled densely enough to approximate the underlying continuous curve, the Varifold Moment Invariants (VMI) exhibit strong robustness, yielding discriminative features for the classification task.
    \item The optimal number of sampled points is dataset-dependent. For instance, as shown in Table~\ref{tab:parametrization}, datasets in which shape differences are primarily driven by size related features (e.g., the HeLa Kyoto dataset \cite{held2010cellcognition}) benefit from lower resolutions than those dominated by small geometric details. For datasets of leaves, (e.g. Flavia \cite{wu2007leaf}), the boundary contain class-discriminating features that are capture with more accuracy with more points along the contours, leading to a slightly improvement of classification results with the increase of the resolution. 
    \item The computational cost of VMI extraction scales linearly with the number of contour points, suggesting an overall complexity of $\mathcal{O}(N)$.
\end{itemize}

\begin{table}[ht]
\centering
\caption{Varifold Moment Invariants computed across different contour parameterizations. The table shows how varying the number of sampled points affects the $F_1$ score classification score (higher is better), as well as how the computational time increases approximately linearly.}
\label{tab:parametrization}
\begin{tabular}{lccc}
\hline
Dataset & $N_{\mathrm{points}}$ & $F_1$-Score & Time (s) \\
\hline
Flavia \cite{wu2007leaf}& 300 & $ 0.9551 \pm 0.0085 $ & $ 347 \textrm { sec }$ \\
Flavia \cite{wu2007leaf}& 400 & $ 0.9578 \pm 0.0081 $ & $ 467 \textrm { sec }$ \\
Flavia \cite{wu2007leaf}& 500 & $ 0.9595 \pm 0.0103 $ & $ 573 \textrm { sec }$ \\
\midrule
MPEG-7 \cite{MPEG-7}& 300 & $ 0.9326 \pm 0.0220 $ & $ 253 \textrm { sec }$ \\
MPEG-7 \cite{MPEG-7}& 400 & $ 0.9265 \pm 0.0187 $ & $ 327 \textrm { sec }$ \\
MPEG-7 \cite{MPEG-7}& 500 & $ 0.9290 \pm 0.0252 $ & $ 402\textrm { sec }$ \\
\midrule
HeLa Kyoto \cite{held2010cellcognition}& 300 & $  0.9359 \pm 0.0379 $ & $ 57\textrm { sec }$ \\
HeLa Kyoto \cite{held2010cellcognition}& 400 & $ 0.9136 \pm 0.0190 $ & $ 75\textrm { sec }$ \\
HeLa Kyoto \cite{held2010cellcognition}& 500 & $ 0.9278 \pm 0.0350 $ & $ 91\textrm { sec }$ \\
\hline
\end{tabular}
\end{table}

\end{document}